\documentclass[10pt,journal,compsoc]{IEEEtran}
\ifCLASSOPTIONcompsoc
  \usepackage[nocompress]{cite}
\else
  \usepackage{cite}
\fi

\usepackage{cite}
\usepackage{graphicx}
\usepackage{textcomp}
\usepackage{xcolor}
\def\BibTeX{{\rm B\kern-.05em{\sc i\kern-.025em b}\kern-.08em
    T\kern-.1667em\lower.7ex\hbox{E}\kern-.125emX}}
    
\usepackage{algorithm}
\usepackage{algpseudocode}
\usepackage{booktabs}
\usepackage{setspace}
\usepackage{subfigure}
\usepackage{amsmath}

\usepackage{multirow}
\usepackage{amssymb}
\usepackage{amsfonts}
\usepackage{mathrsfs}

\newcommand{\argmaxF}{\mathop{\mathrm{argmax}}\limits}

\newcommand{\argminF}{\mathop{\mathrm{argmin}}\limits}
\newtheorem{prop}{Proposition}
\usepackage{amssymb}% http://ctan.org/pkg/amssymb
\usepackage{pifont}% http://ctan.org/pkg/pifont

\usepackage{soul} % newly add
\usepackage{tabularx} % newly add
\usepackage{tabstackengine} % newly add
\usepackage{url} % newly add for citing an url

\usepackage{color}

\newcommand{\para}[1]{\vspace{1mm}\noindent\textbf{#1.}}

\usepackage{tikz}
\newcommand*\circled[1]{\tikz[baseline=(char.base)]{
            \node[shape=circle,draw,inner sep=0.8pt] (char) {#1};}}

\def\NoNumber#1{{\def\alglinenumber##1{}\State #1}\addtocounter{ALG@line}{-1}}
\ifCLASSINFOpdf
\else
\fi
\begin{document}
\title{Privacy-preserving Federated Adversarial Domain Adaptation over Feature Groups for Interpretability}

\author{\IEEEauthorblockN{Yan Kang\IEEEauthorrefmark{1},
Yuanqin He, 
Jiahuan Luo, 
Tao Fan,
Yang Liu,
and Qiang Yang}

\thanks{Corresponding author: Yan Kang (email: yangkang@webank.com).}
}

\markboth{Journal of \LaTeX\ Class Files,~Vol.~14, No.~8, August~2022}%
{Shell \MakeLowercase{\textit{et al.}}: Bare Demo of IEEEtran.cls for Computer Society Journals}



\IEEEtitleabstractindextext{%
\begin{abstract}
We present a novel privacy-preserving federated adversarial domain adaptation approach ($\textbf{PrADA}$) to address an under-studied but practical cross-silo federated domain adaptation problem, in which the party of the target domain is insufficient in both samples and features. We handle the lack-of-feature issue by extending the feature space through vertical federated learning with a feature-rich party and tackle the sample-scarce issue by performing adversarial domain adaptation from the sample-rich source party to the target party. In this work, we focus on financial applications where interpretability is critical. However, existing adversarial domain adaptation methods typically apply a single feature extractor to learn low-interpretable feature representations with respect to the target task. To improve interpretability, we exploit domain expertise to categorize the feature space into multiple groups that each group holds tightly relevant features, and we learn a semantically meaningful high-order feature from each feature group. In addition, we apply a fine-grained domain adaptation to each feature group to improve transferability. We design a privacy-preserving vertical federated learning framework that enables performing the PrADA securely and efficiently. We evaluate our approach based on two tabular datasets. Experiments demonstrate both the effectiveness and practicality of our approach.
\end{abstract}

}

\IEEEpubid{\scriptsize Copyright (c) 2022 IEEE. Personal use is permitted. For any other purposes, permission must be obtained from the IEEE by emailing pubs-permissions@ieee.org.}
\maketitle

\IEEEdisplaynontitleabstractindextext

% \fi
\IEEEpeerreviewmaketitle

\IEEEraisesectionheading{\section{Introduction}\label{sec:introduction}}

\IEEEPARstart{D}{omain} adaptation approaches~\cite{pmlr-v37-ganin15,Tzeng2017,Saito_2018_CVPR,Wang2019,pmlr-v97-zhang19i} have shown notable success. Those approaches typically establish alignment or minimize the discrepancy between source and target domains by creating domain-invariant feature representation in the form of deep neural network (DNN) feature extractors. In addition to the remarkable ability of DNN on encoding raw data into meaningful representations that result in high performance on objective tasks, a major enabler of the adoption of DNN in domain adaptation is the availability of a large amount of data with rich features (image and text) that supports the representation learning of DNN. 

% However, in many real-world federated learning scenarios where the data is in tabular format (i.e., sample-feature matrix), the participating parties might be insufficient in features. Thus, DNN-based feature extractors would be overparametrized and thus fail to adapt to the target domain. 

% However, in many real-world federated learning applications where the tabular data (i.e., sample-feature matrix) is the mainstream, the participating parties might be insufficient in features to support domain adaptation. One promising way to address the lack-of-feature issue is to enlarge the feature space by collaborating with a feature-rich party. For example, financial institutes with limited features (e.g., personal credit report) may have a large number of overlapping users with an e-commerce company that curates rich user information (e.g., product-browsing history  and app usage information) and therefore they can collaboratively build domain adaptation models based on the enlarged feature space. 

Due to increasingly strict legal and regulatory constraints enforced on user privacy, private data from different organizations (domains) cannot be directly integrated for training machine learning models. In recent years, federated learning (FL) has emerged as a practicable solution to tackle data silo issues without compromising user privacy. Initially, FL \cite{McMahan2017Communication} was proposed to build models by utilizing data of millions of mobile devices. \cite{Yang2019Federated} further extends FL architecture to enterprise setting where participating parties might be much smaller but privacy concerns are paramount. This setting is coined as \textit{cross-silo federated learning} \cite{kairouz2019federated}.

Recently, a growing number of works have been proposed to integrate domain adaptation into cross-silo FL setting ~\cite{Peterson2019PrivateFL,peng2019federated,LI2020101765,song9157838} for solving domain shift issues among independent parties. These federated domain adaptation (FDA) methods conduct experiments typically using image and text data that have rich features to perform meaningful representation learning. However, in many real-world FL applications where data is stored in tabular format (i.e., sample-feature matrix), the participating parties might be insufficient in features for building DNN-based domain adaptation models. One promising way to address the lack-of-feature issue is to enlarge the feature space by collaborating with a feature-rich party. For example, financial institutes with limited features (e.g., only basic user info) may have a large number of overlapping users with an e-commerce site that curates rich user information (e.g., product-browsing history and app usage information) and thereby they can collaboratively build domain adaptation models based on the enlarged feature space. This cross-silo FL setting where sample features distributed in different parties is categorized as \textit{vertical (feature-partitioned) federated learning (VFL)} \cite{Yang2019Federated}.

Although enlarged feature space enables domain adaptation, mainstream adversarial DA methods~\cite{pmlr-v37-ganin15,Tzeng2017,Wang2019} typically apply one single pair of feature extractor and domain discriminator over the whole feature space to learn feature representations, which is not understandable by human. In this work, we focus on financial applications in which the model interpretability is an important concern. Thus, training models directly on top of raw feature representations cannot satisfy our requirements toward model interpretability and regulation. In addition, a single pair of feature extractor and domain discriminator may not be effective to learn transferable feature representations. In this work, we propose to group highly relevant features together and apply domain adaptation to each feature group aiming to improve both interpretability and transferability.

% Although leveraging rich features from a third party is a promising way to improve the performance of the target model, those features are typically high-dimensional, and a large portion of them are likely to be low-predictive and low-interpretable with respect to the target task. In this work, we focus on financial applications in which the model interpretability is an important concern. Thus, training the target model directly on top of those raw features cannot satisfy our requirements toward model interpretability and regulation. In addition, the mainstream adversarial DA~\cite{pmlr-v37-ganin15,Tzeng2017,Saito_2018_CVPR,Wang2019,pmlr-v97-zhang19i} of applying a single pair of feature extractor and domain discriminator over the whole feature space to learn transferable feature representations may not effectively disentangle different explanatory factors of tabular features. We propose to split the feature space into groups of relevant features and apply fine-grained DA to each group to improve both feature transferability and model interpretability. 

Most FDA approaches apply differential privacy (DP)\cite{dwork2006calibrating,dwork2011firm,dwork2014algorithmic} to protect the privacy of participants' private data. But DP suffers from precision loss, which is not acceptable in high-stake decision-making applications (e.g., financial services and healthcare) where precision is crucial. 

In this work, we propose \textbf{PrADA}, a privacy-preserving federated adversarial domain adaptation approach that enables participating parties to collaboratively conduct domain adaptation modeling in a privacy-preserving manner while taking the model interpretability into account.
% In addition to precision, model explainability is another important concern in high-cost decision-making applications for supporting explanatory, debugging and regulatory purposes.
% Tabular data is the most common data format in real-world machine learning applications \cite{Michael2018} ranging from retail, finance to E-commerce.
The main contributions of this work are highlighted as follows: 
\begin{enumerate}
    \item To our best knowledge, this work is the first study on domain adaptation problem in the VFL setting for tabular data;
    % When tabular data is limited in features, DNN-based feature extractor may be overparameterized. Through VFL, we take full advantage of sample features located in different parties to support domain adaptation modeling;
    \item This work proposes a fine-grained adversarial domain adaptation approach to reduce feature dimensionality, enhance model interpretability, and facilitate the learning of domain-invariant features.
    % (section \ref{fada}).
    \item This work proposes a privacy-preserving VFL framework that allows participating parties to collaboratively conduct domain adaptation without exposing private local data under the \textit{semi-honest} assumption.
    % (section \ref{sfl}).
    % and we provide security analysis proving that no data leakage would occur under the semi-honest assumption (section \ref{sfl});
\end{enumerate}

% and (iii) model interpretablility is an important concern, much simpler models such as logistic regression and variants of ensemble decision trees are dominant while DNN is under-explored. The reasons for this are straightforward. DNN requires a large amount of training samples with rich features and decisions it made are hard to explain.

% Collecting a large amount of training samples with rich features is expensive. DNN are overparametrized for small data with very limited feature.

% they are highly interpretable in their basic form (e.g. by tracking decision nodes) and there are effective post-hoc explainability methods for their ensemble form, e.g. [36] – this is an important concern in many real-world applications (e.g. in financial services, where trust behind a high-risk action is crucial); 

\section{Related Work}

\subsection{Federated Domain Adaptation}
Traditional domain adaptation (DA) approaches assume the data are  centralized on one server, thus limiting their applicability to decentralized real-world scenarios. Federated domain adaptation aims to conduct domain adaptation modeling among independent parties of different domains without violating privacy. 
\cite{Peterson2019PrivateFL} applies a mixture of experts (MoE) strategy that each participant combines a collaboratively-learned general model and a domain-tuned private model to reconcile distribution differences among participants. \cite{peng2019federated} leverages federated adversarial domain alignment with a dynamic attention mechanism to enhance knowledge transfer. \cite{LI2020101765} applies methods proposed by \cite{Peterson2019PrivateFL,peng2019federated} to functional magnetic resonance imaging (fMRI) analysis. \cite{7e114b6ca0744925b9974516d3dfbdb6} proposes agnostic federated learning aiming to optimize the global model for any target distribution formed by a mixture of client distributions without overfitting data of any particular client. One major limitation of those (both traditional and federated ) DA approaches is that they almost use computer vision datasets, and only a few of them (e.g., \cite{7e114b6ca0744925b9974516d3dfbdb6}) are evaluated on tabular data. 

% Most traditional and federated domain adaptation applications are for computer vision. 

% \cite{song9157838} extends the secure federate transfer learning framework proposed by \cite{LiuKang20209076003} to tackle federated unsupervised domain adaptation problem. It enables the source and target domains to minimize the squared maximum mean discrepancy (MMD) loss in a distributed manner and protect data privacy using PHE. 

\subsection{Deep Neural Network on Encrypted Data}

Protecting privacy is a crucial element of federated learning. Homomorphic encryption (HE) is one of the major solutions to address the privacy issue. Although HE is a promising solution that allows computation to be performed on encrypted data, its expensive computational cost makes it impractical to be applied in training an entire DNN model. To address this issue, GELU-NET~\cite{ijcai2018-547} adopts a client-server architecture in which the client encrypts the data while the server performs most computation on encrypted data. ACML~\cite{Zhang2020AdditivelyHE} focuses on a more enterprise scenario where data and labels are distributed among two independent parties. It adopts a SplitNN~\cite{vepakomma2018peek} approach that each party is only responsible for updating its own portion of the whole DNN model. The novelty of ACML is that the costly encryption-decryption operations are only performed on the two partial models' boundaries, leaving the rest of the computation in plaintext.

\subsection{Model Interpretability}

A variety of research works have been proposed for interpreting deep neural networks \cite{Montavon2017}. These methods focus on post-hoc interpretability that analyzes the relationships between input and output of the trained model rather than elucidating models' internal structures. Other methods \cite{NEURIPS2019_adf7ee2d,Daniel2020,Alvarez2018} construct prototypes or general concepts that shed light on the decision-making process. e.g.,  \cite{NEURIPS2019_adf7ee2d} propose ProtoPNet that learns a set of prototypes each can be considered as the latent representation of a small prototypical part of training images. Then, the label prediction can be calculated based on a weighted combination of the similarity scores between parts of the image and the learned prototypes. \cite{NIPS2017_7062} calculates SHAP values of every feature for every sample based on model prediction. Complex models, such as ensemble methods or deep networks, can be explained through these SHAP values.

\section{Problem Definition}

We consider following cross-silo federated domain adaptation scenario that involves three parties. Party A is from the \textit{target} domain, and it has a small number of labeled samples $(\mathbf{X}^{A}_{l}, \mathbf{Y}^{A}) \in \mathbb{R}^{n^A_l \times (m+1)}$ and some unlabeled samples $\mathbf{X}^{A}_{u} \in \mathbb{R}^{n^A_u \times m}$. Party B is from the \textit{source} domain and it has a large amount of labeled samples $(\mathbf{X}^{B}, \mathbf{Y}^{B}) \in \mathbb{R}^{n^B \times (m+1)}$. $n^A = n^A_l + n^A_u$ and $n^B$ denote the sample size of parties A and B respectively, while $m$ denotes the feature dimension. These two parties share the same feature space and have similar tasks. We consider conduct domain adaptation (DA) from party B to party A, and we call these two parties \textit{active parties} because they initiate the DA procedure. The two active parties have insufficient number of features to support DA. Thus, we refer to a \textit{passive party} C that is able to provide sufficient amount of complementary features $\mathbf{X}^{B^c} \in \mathbb{R}^{n^B \times m^c}$ and $\mathbf{X}^{A^c} \in \mathbb{R}^{n^A \times m^c}$ for party B and party A, respectively. $\mathbf{X}^{B^c}$ and $\mathbf{X}^{A^c}$ have the same feature space with dimension $m^c$, and $n^B \gg n^A_l$ and $m^c \gg m$. 
 
We align $\mathbf{X}^{A^c}$ with $(\mathbf{X}^{A}_l, \mathbf{Y}^{A})$ and $\mathbf{X}^{A}_u$ respectively along the feature axis to form a virtual labeled dataset $\mathbf{D}^t_l = [\mathbf{X}^{A^c}_l;\mathbf{X}^{A}_l;\mathbf{Y}^{A}]$ and a virtual unlabeled dataset $\mathbf{D}^t_u = [\mathbf{X}^{A^c}_u;\mathbf{X}^{A}_u]$ of the target domain. Likewise, we form a virtual dataset $\mathbf{D}^s = [\mathbf{X}^{B^c};\mathbf{X}^{B};\mathbf{Y}^{B}]$ of the source domain. The alignment can be performed by leveraging privacy-preserving entity matching approaches \cite{Hardy2017}. Figure 1 shows the federated view of tabular datasets $\mathbf{D}^s$ and $\mathbf{D}^t$ among the three parties.

\begin{figure}[ht]
\centering
\includegraphics[width=0.99\linewidth]{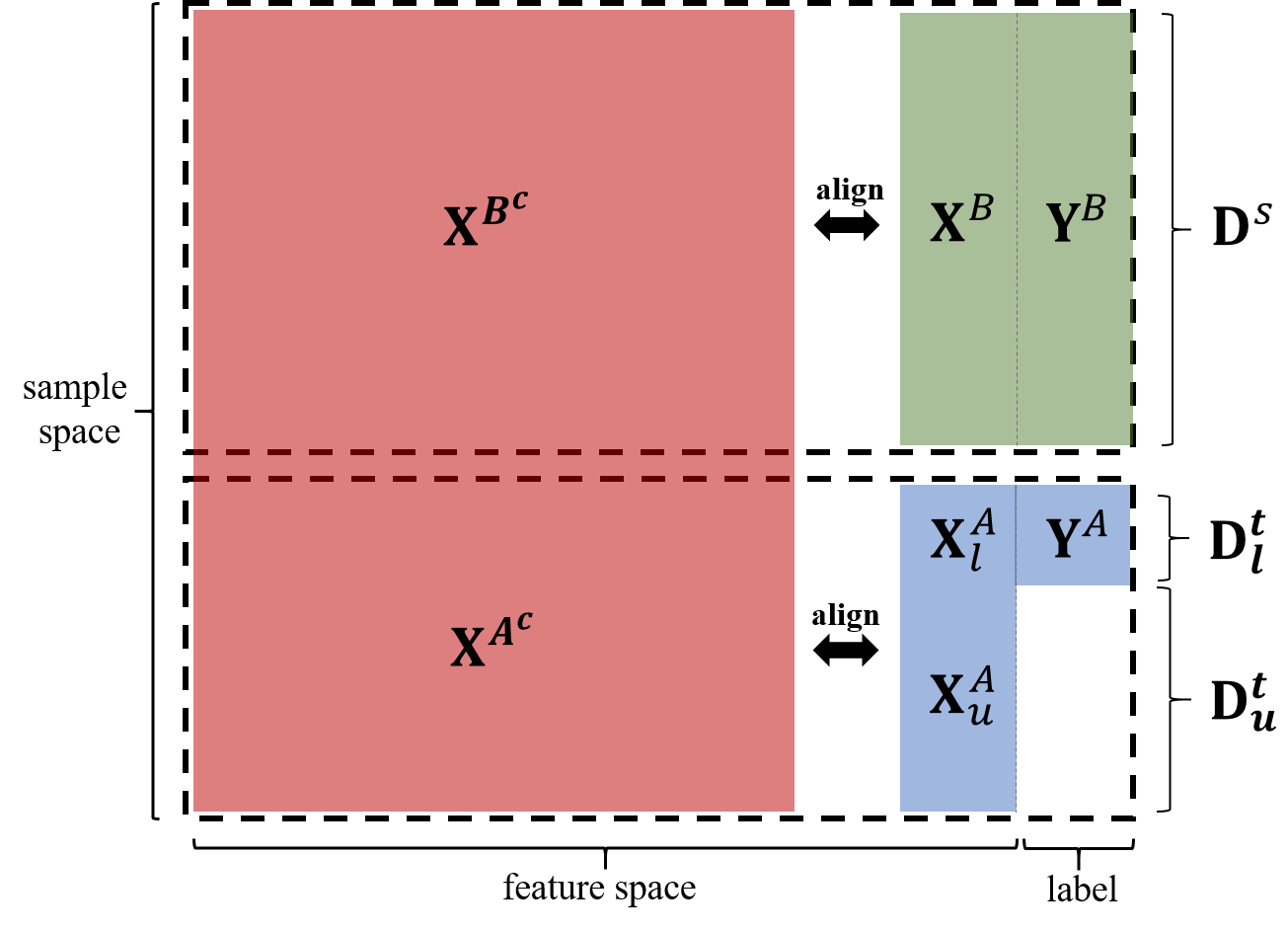}
\caption{View of the virtual tabular data of the cross-silo federated domain adaptation. Source party B has a large amount of labeled samples $(\mathbf{X}^{B}, \mathbf{Y}^{B})$, while target party A has a small amount of labeled samples $(\mathbf{X}^{A}_l, \mathbf{Y}^{A})$ and some unlabeled samples $\mathbf{X}^{A}_u$. Party C provides complementary features $\mathbf{X}^{A^c}$ for party A and $\mathbf{X}^{B^c}$ for party B. Thus, we form virtual dataset $\mathbf{D}^s = [\mathbf{X}^{B^c};\mathbf{X}^{B};\mathbf{Y}^{B}]$ of the source domain, and virtual datasets $\mathbf{D}^t_l = [\mathbf{X}^{A^c}_l;\mathbf{X}^{A}_l;\mathbf{Y}^{A}]$ and $\mathbf{D}^t_u = [\mathbf{X}^{A^c}_u;\mathbf{X}^{A}_u]$ of the target domain.} 
\label{IFTL_data}
\end{figure}

% Since the features and labels are provided by two different parties for both domains, the supervised domain adaptation needs to be performed in federated learning manner where privacy-preserving protocol is applied. Figure 1 illustrates the tabular view of $\mathbf{D}^s$ and $\mathbf{D}^t$

% In this work, we refer to party A and party B as \textit{active parties} because they initiate the domain adaptation learning procedure, while party C as the \textit{passive party}. 

Under this setting, our PrADA approach is conducted from two aspects: (1) extending the feature space of active parties A and B through vertical federated learning with a feature-rich passive party C; (2) performing domain adaptation from party B of the sample-rich source domain to party A of the sample-scarce target domain based on the extended yet distributed feature space. \textit{Our ultimate goal is to improve the performance of the target model of party A}.

Because $\mathbf{D}^s$ and $\mathbf{D}^t$ are composed of data from two independent parties, this domain adaptation is performed in a federated learning manner where a privacy-preserving protocol is applied. We assume that all three parties are honest-but-curious, meaning they follow the federated learning protocol but attempt to deduce as much as possible from the information received from other parties. 

\section{Architecture Overview}
\begin{figure*}
	\centering
	\includegraphics[width=0.99\linewidth]{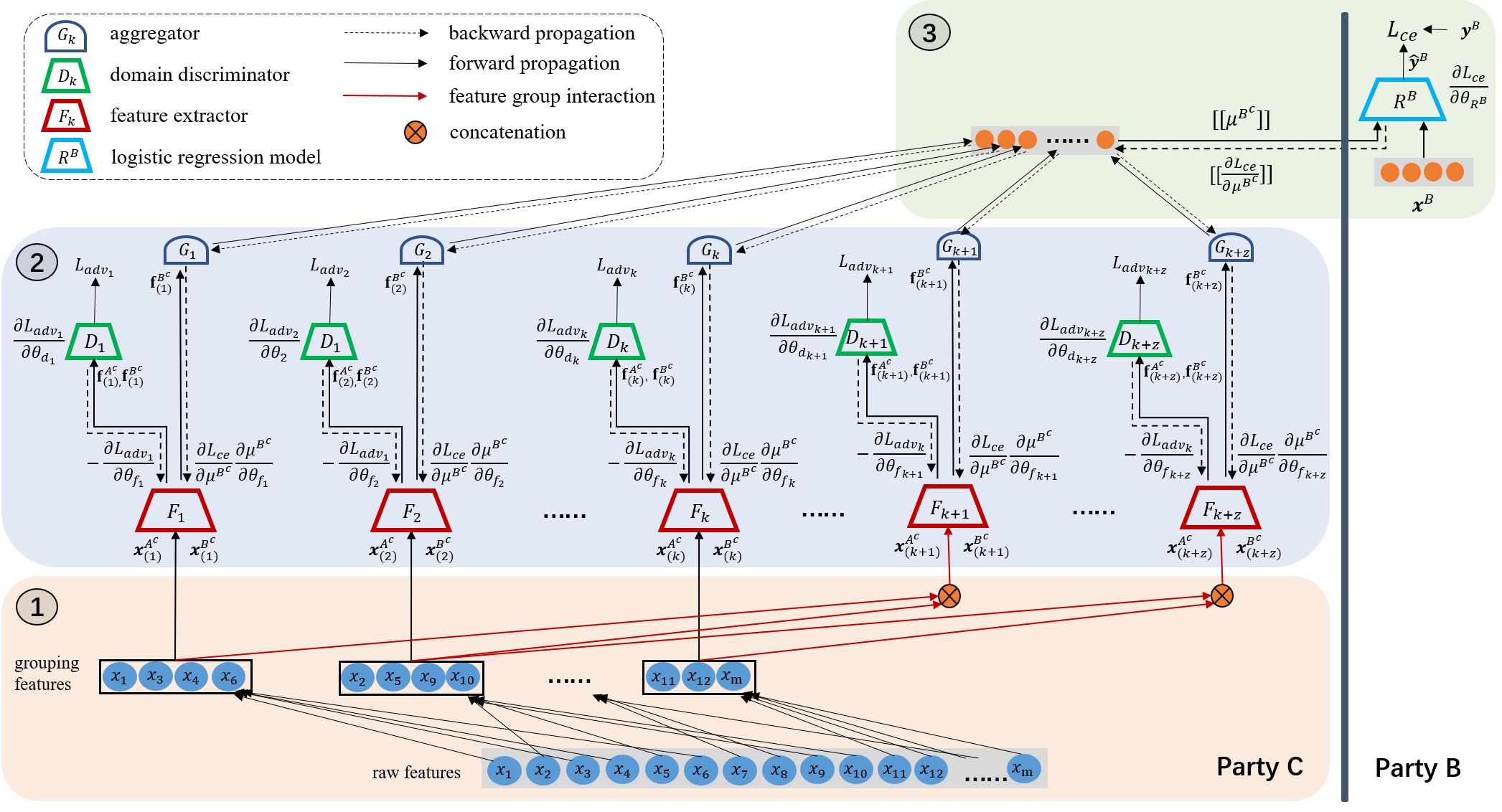}
	\caption{The pre-training stage of PrADA. 
	$\theta_{f_i}, \theta_{d_i}, \theta_{R^B}$ denote model parameters of $i$th feature extractor, $i$th  discriminator and label predictor $R^B$, respectively. Party C locally trains $g$ pairs of $\{(\theta_{f_i}, \theta_{d_i})\}_{i=1}^{g}$ that each corresponds to a feature group by optimizing (\ref{argminmax_F_D}). Party B and party C collaboratively train $\{\theta_{f_i}\}_{i=1}^{g}$ and $\theta_{R^B}$ by optimizing (\ref{argmin_F_R}) using $\mathbf{\mu}^{B^c}$, $\mathbf{x}^{B}$ and $y^B$.}
	\label{feddann}
\end{figure*}

PrADA involves two stages: \textit{pre-training} and  \textit{fine-tuning}. The pre-training stage is performed between source party B and party C aiming to pre-train feature extractors maintained by party C, while fine-tuning is performed between target party A and party C, aiming to train the target label predictor of party A based on pre-trained feature extractors. 
Figure \ref{feddann} illustrates the workflow of the pre-training stage of the federated adversarial domain adaptation. Since the solely goal of fine-tuning is to train the target label predictor of party A, fine-tuning follows a similar workflow as pre-training except that no domain adaptation is involved. 
% The fine-tuning stage has a similar workflow except that no domain adaptation is involved. 
% Figure \ref{feddann} illustrates the architecture of models in the pre-training stage. The fine-tuning stage has a similar architecture except that no domain adaptation is involved. 

% Figure \ref{feddann} shows that party B owns label predictor $R^B$ while party C owns feature extractors $\mathscr{F} = \{F_i\}_{i=1}^g$ and domain discriminators $\mathscr{D} = \{D_i\}_{i=1}^g$ for extracting transferable feature representations from multiple feature groups. Instead of directly passing feature representations to party B for training the label predictor $R^B$, party C leverages a set of aggregators $\mathscr{G} = \{G_i\}_{i=1}^g$ to compress each feature representation learned from a feature group into a scalar value representing a high-order feature and feeds these high-order features into the label predictor. Party B has no feature extractors because we assume that party B has insufficient raw features to conduct effective domain adaptation, and more importantly, these raw features (e.g., age, gender, education) are business-related such that their contributions to the model decision should be interpretable. Therefore, party B feeds their raw features directly into their label predictors. The feature extractors and domain discriminators are trained locally at party C, while the label predictor is trained collaboratively by both party B and party C in a privacy-preserving manner, which we will elaborate in the section \ref{sfl}. 

As illustrated in Figure \ref{feddann}, The party B owns label predictor $R^B$ while party C owns feature extractors $\mathscr{F} = \{F_i\}_{i=1}^g$ and their corresponding domain discriminators $\mathscr{D} = \{D_i\}_{i=1}^g$, and aggregators $\mathscr{G} = \{G_i\}_{i=1}^g$. The pre-train stage of federated adversarial domain adaptation mainly involves three steps. 

\para{\circled{1} Feature grouping} The party C leverages domain expertise to group raw features into $k$ feature groups that each comprises tightly relevant features. In addition, the party C forms $z$ interactions between pairwise feature groups. Thus, this step gives totall $g=k+z$ feature groups. ($k$ normal feature groups and $z$ interactive feature groups). 

\para{\circled{2} Adversarial domain adaptation} The party C leverages the adversarial domain adaptation to train feature extractors $\mathscr{F} = \{F_i\}_{i=1}^g$ in order to learn domain-invariant feature representations based upon those $g$ feature groups.

\para{\circled{3} Vertical federated learning} The party B and party C collaboratively perform vertical federated learning to train the task-specific label predictor $R^B$ and feature extractors $\mathscr{F} = \{F_i\}_{i=1}^g$ in order to learn domain-specific feature representations.

We discuss \circled{1} in section \ref{feature_grouping} and elaborate \circled{2} \circled{3} in section \ref{fada}. In section \ref{sfl}, we explain how our privacy-preserving vertial federated learning framework is applied to protect data privacy of the whole workflow.

% Instead of directly passing feature representations to party B for training the label predictor $R^B$, party C leverages a set of aggregators $\mathscr{G} = \{G_i\}_{i=1}^g$ to compress each feature representation learned from a feature group into a scalar value representing a high-order feature and feeds these high-order features into the label predictor. Party B has no feature extractors because we assume that party B has insufficient raw features to conduct effective domain adaptation, and more importantly, these raw features (e.g., age, gender, education) are business-related such that their contributions to the model decision should be interpretable. Therefore, party B feeds their raw features directly into their label predictors. The feature extractors and domain discriminators are trained locally at party C, while the label predictor is trained collaboratively by both party B and party C in a privacy-preserving manner, which we will elaborate in the section \ref{sfl}. 

% The party C trains its feature extractors and domain discriminators locally while the label predictor 

\section{Feature Grouping} 
\label{feature_grouping}
% We consider both active parties adopt logistic regression model (LR) as the label predictor because LR model is explainable \cite{molnar2019} and it can be trained end-to-end with neural networks.
% Different from mainstream adversarial DA methods \cite{pmlr-v37-ganin15,Tzeng2017,Wang2019} that align the source domain to the target domain typically by training a single feature extractor to learn domain-invariant feature representations of the entire feature space, 
% PrADA leverages \textit{feature grouping} to improve both the transferability of feature extractors and the interpretabiltiy of label predictors. We propose that the party C creates $k$ feature groups out of its original feature space such that features in the same group are more relevant than features belonging to other groups. To further improve model performance, party C performs interaction between each pair of these $k$ feature groups giving $z$ \textit{interactive} feature groups. As a result, party C creates $k+z$ feature groups. Naturally, party C assigns each feature group a feature extractor along with a domain discriminator to learn domain-invariant features representations. We hypothesize that a more fine-grained adaptation between each of the two domains' feature group that includes tightly relevant features helps improve the transferability of domain-invariant feature representations.

The reasons that the PrADA leverages \textit{feature grouping} are two folds: (1) improve the transferability of feature extractors; (2) improve the interpretabiltiy of label predictors. 

We propose that with the help of domain expertise the party C creates $k$ feature groups out of its original feature space such that features in the same group are more relevant than features belonging to other groups. Based on feature grouping, the party C obtains $k$ groups of relevant features $\{\mathbf{x}^{p^c}_{(i)}\}_{i=1}^{k}$ for each sample $\mathbf{x}^{p^c} \in \mathbb{R}^{1\times m^c}$ dawn from $\mathbf{X}^{p^c}, p \in \{A,B\}$. To explore interactive features, party C performs interaction between each pair of the $k$ feature groups by concatenating the two feature groups together, giving $z=C_2^k$ interactive feature groups. As a result, the party C creates totally $g=k+z$ feature groups. Naturally, the party C assigns each feature group a feature extractor along with a domain discriminator to learn domain-invariant features representations. We hypothesize that this fine-grained domain adaptation between each of the two domains' feature group that includes tightly relevant features helps improve the transferability of domain-invariant feature representations.

% To further improve model performance, party C performs interaction between each pair of these $k$ feature groups giving $z$ \textit{interactive} feature groups. As a result, party C creates $k+z$ feature groups. Naturally, party C assigns each feature group a feature extractor along with a domain discriminator to learn domain-invariant features representations. We hypothesize that a more fine-grained adaptation between each of the two domains' feature group that includes tightly relevant features helps improve the transferability of domain-invariant feature representations.

We adopt logistic regression (LR) model as the label predictor because it is a widely used interpretable model in financial applications. LR considers each feature and its associated weight as a fundamental interpretable unit. Therefore, instead of directly passing dense feature representations output from feature extractors to the LR model, party C leverages a set of aggregators $\{G_i\}_{i=1}^g$ that compress the output of each feature extractor into a scalar value representing a high-order feature, and then it feeds these high-order features into the LR model. As a result, the LR model takes as input a manageable number of high-order features (from party C), which are more explainable than it takes as input the concatenation of multiple dense feature representations. We formalize the procedure of party C generating high-order features vector $\boldsymbol\mu^{p^c}$ as follows:
\begin{equation} 
\label{mu_C}
% g(\mathbf{x}^{s,C}) 
\boldsymbol\mu^{p^c} = [G_1(\mathbf{f}^{p^c}_{(1)});\dots;G_k(\mathbf{f}^{p^c}_{(k)});\dots;G_g(\mathbf{f}^{p^c}_{(g)})]
\end{equation}
where $\mathbf{f}_{(i)}^{p^c}$ denotes feature representations learned from feature extractor $F_i(\mathbf{x}^{p^c}_{(i)})$, and $G_i(\mathbf{f}^{p^c}_{(i)})$ returns a scalar value representing the high order feature for feature group $\mathbf{x}_{(i)}^{p^c}$.

For performing federated adversarial domain adaptation, the party C feeds $\{\mathbf{f}_{(i)}^{p^c}\}_{i=1}^{g}$ into their corresponding domain discriminators for optimizing domain discrimination losses and passes high-order feature vectors $\boldsymbol\mu^{p^c}$ to active party $p$ (together with $p$'s own raw features) for optimizing label prediction loss, as described in section \ref{fada}. 

\section{Federated Adversarial Domain Adaptation}
\label{fada}

% The core idea of adversarial domain adaptation is to learn a feature extractor that is able to extract features that are both discriminative to the task and invariant to the change of domains. This typically is achieved by jointly optimizing a domain discriminator that tries to tell apart source and target domains, and a label predictor that predicts class labels [ref].

% The essential idea of adversarial domain adaptation is to train a feature extractor that is able to learn features that are both discriminative to the task and invariant to the change of domains such that the label predictor trained on data of source domain can work well on data of target domain [ref].

Federated adversarial domain adaptation of PrADA involves two stages:  \textit{pre-training} and  \textit{fine-tuning}. The pre-training stage is performed collaboratively between source party B and party C, and it aims to train feature extractors that can learn both domain-invariant and label-discriminative features. The fine-tuning stage is performed collaboratively between target party A and party C and it aims to train the target label predictor possessed by party A leveraging pre-trained feature extractors.

\subsection{Pre-training Stage}
The essential idea of adversarial domain adaptation is to train feature extractors that are able to learn features that are both discriminative to the task and invariant to the change of domains. Thus, we have two optimization goals. (1) In order to obtain domain-invariant features, we perform adversarial domain adaptation to optimize the feature extractors that maximizes the domain classification losses, while simultaneously optimize the domain discriminators that minimizes the domain classification loss. (2) To obtain task-specific discriminative feature representations, we perform vertical federated learning (VFL) to optimize the feature extractors and label predictor that minimize the label prediction loss. 

% In our federated learning setting, party C curates both source data and target data and thus it is responsible for optimizing domain classification losses and train relevant models. Unlike the traditional adversarial domain adaptation approaches [ref, ref, ref] that typically utilize one feature extractor and one domain discriminator to learn domain-invariant feature representations, party C leverage $k$ feature extractors $\mathscr{F} = \{F_i\}_{i=1}^k$ and their corresponding $k$ domain discriminators $\mathscr{D} = \{D_i\}_{i=1}^k$ to learn domain-invariant feature representations from $k$ feature groups. More specifically, the $i$th feature extractor $F_i$ learns domain-invariant features from the $i$th feature group and the output of $F_i$ is mapped to the domain label $d \in \{0, 1\}$ by the $i$th domain discriminator $D_{i}$. The overall domain classification loss is the sum of domain classification loss for each domain discriminator $D_{i}$. We give the formula as follows:
In our federated learning setting, party C leverages $g$ feature extractors $\mathscr{F} = \{F_i\}_{i=1}^g$ and their corresponding $g$ domain discriminators $\mathscr{D} = \{D_i\}_{i=1}^g$ to learn domain-invariant feature representations from $g$ feature groups. More specifically, the $i$th feature extractor $F_i$ learns feature representation from the $i$th feature group and then the $i$th domain discriminator $D_{i}$ maps this feature representation to a domain label $d \in \{0, 1\}$. The overall domain classification loss is the sum of domain classification losses for all domain discriminators in $\mathscr{D}$. We give the formula as follows:
\begin{multline}
\label{loss_adv}
L_{adv}(\mathscr{F}, \mathscr{D}) = - \mathbb{E}_{\mathbf{x}^{A^c} \sim \mathbf{X}^{A^c}} \sum_{i=1}^{g}log[D_i(F_i(\mathbf{x}_{(i)}^{A^c}))] \\
- \mathbb{E}_{\mathbf{x}^{B^c} \sim \mathbf{X}^{B^c}} \sum_{i=1}^{g}log[1 - D_i(F_i(\mathbf{x}_{(i)}^{B^c}))]
\end{multline}
To make feature extractors obtain task-specific discriminative features, we optimize label prediction loss to train both label predictor and feature extractors to classify the source samples correctly. We define the label prediction loss as:
\begin{multline}
\label{loss_bce}
L_{ce}(\mathscr{F}, R^B) = \\ \mathbb{E}_{(\mathbf{x}^{B^c}, \mathbf{x}^{B}, \mathbf{y}^{B}) \sim \mathbf{D}^{s}} [\ell_{ce}(R^B([\boldsymbol\mu^{B^c};\mathbf{x}^{B}]), \mathbf{y}^{B})]
\end{multline}
where $R^B$ is the label predictor curated at party B, $\boldsymbol\mu^{B^c}$ is the high-order feature vectors passed from party C and $\mathbf{x}^{B} $ is the feature vectors possessed by party B. 

The pre-training stage optimizes the two losses presented in (\ref{loss_adv}) and (\ref{loss_bce}). We have the complete loss function for pre-training stage as follows:
% Combining the two losses presented in ep (\ref{loss_adv}) and ep (\ref{loss_bce}), we have the complete optimization objective for pre-training as follow:
\begin{equation}
L(\mathscr{F}, \mathscr{D}, R^B) = L_{ce}(\mathscr{F}, R^B) - \lambda L_{adv}(\mathscr{F}, \mathscr{D})
\end{equation}
where $\lambda$ is a hyperparameter that controls the trade-off between the two losses that shape the feature representations during training. 
% $L_{adv}$ and $L_{ce}$ are optimized separately: 

In our federated setting, $L_{adv}(\mathscr{F}, \mathscr{D})$ is optimized locally at party C since it only involves data of party C, while $L_{ce}(\mathscr{F}, R^B)$ is optimized collaboratively by party B and party C in a federated manner since it involves data from the two parties. To this end, we train parameters $\{\theta_{f_i}\}_{i=1}^g$ of feature extractors $\mathscr{F}$, $\{\theta_{d_i}\}_{i=1}^g$ of domain discriminators $\mathscr{D}$, and $\theta_{R^B}$ of label predictor $R^B$ by solving following two optimization problems.
\begin{equation}\label{argminmax_F_D}
\argminF_{\{\theta_{f_i}\}_{i=1}^g} \argmaxF_{\{\theta_{d_i}\}_{i=1}^g} (-\lambda L_{adv}(\mathscr{F}, \mathscr{D}))
\end{equation}
\begin{equation}\label{argmin_F_R}
\argminF_{\{\theta_{f_i}\}_{i=1}^g, \theta_{R^B}} L_{ce}(\mathscr{F}, R^B)
\end{equation}
$\{\theta_{d_i}\}_{i=1}^g$ are trained by minimizing the domain classification loss, $\theta_{R^B}$ is trained by minimizing the label prediction loss, and $\{\theta_{f_i}\}_{i=1}^g$ are trained by minimizing the label prediction loss while at the same time maximizing the domain classification loss.

% In the federated learning setting, $L_{adv}$ and $L_{ce}$ are optimized separately. $L_{adv}(\mathscr{F}, \mathscr{D})$ is optimized locally at party C since it only involves data of party C while $L_{ce}(\mathscr{F}, R^A)$ should be optimized collaboratively by party A and party C in a federated manner since it involves data and models from the two parties. To this end, we rewrite eq (\ref{argminmax_F_R_D}) as following two separate optimization problems. 

% \begin{equation}\label{argminmax_F_D}
% \argminF_{\mathscr{F}} \argmaxF_{\mathscr{D}} (-\lambda L_{adv}(\mathscr{F}, \mathscr{D}))
% \end{equation}

% \begin{equation}\label{argmin_F_R}
% \argminF_{\mathscr{F}, R^A} L_{ce}(\mathscr{F}, R^A)
% \end{equation}

% The overall workflow of pre-training stage is illustrated in Figure \ref{feddann} and the pseudocode for optimizing eq (\ref{argminmax_F_D}) and eq (\ref{argmin_F_R}) is presented in Algorithm \ref{feddann_algo}. 

Figure \ref{feddann} illustrates the overall workflow of the pre-training stage and Algorithm \ref{feddann_algo} describes the procedure of optimizing (\ref{argminmax_F_D}) and (\ref{argmin_F_R}). We assume that the entity alignment procedure has been run and the indices of $\mathbf{D}^s$ have been shuffled and synchronized between party B and party C before training. For each inner iteration, party C and party B fetch the same mini-batch of aligned samples from $\mathbf{D}^s$ but each holds its own portion of private data: party C holds $\mathbf{x}^{B^c}$ while party B holds $\mathbf{x}^{B}$ and $\mathbf{y}^{B}$. In addition, Party C samples a mini-batch $\mathbf{x}^{A^c}$ from its target data $\mathbf{X}^{A^c}$. Based on $\mathbf{x}^{B^c}$ and $\mathbf{x}^{A^c}$, party C optimizes (\ref{argminmax_F_D}) locally. Based on $\mathbf{\mu}^{B^c}$, $\mathbf{x}^{B}$ and $y^B$, party B and party C collaboratively optimize (\ref{argmin_F_R}) through Algorithm 3.

% Next, party C generates feature vectors $\mathbf{\mu}^{A^c}$ from $\mathbf{x}^{A^c}$ applying (\ref{mu_C}). 
%  encrypts these vectors and sends them to party A. 

\begin{algorithm}[h]
\caption{Federated Pre-training}
\label{feddann_algo}
\begin{algorithmic}[1]
% \REQUIRE Model $\mathcal{M}_\mathcal{F}$, number of folds $K$
\State \textbf{Initialization}: feature extractors $\mathscr{F}$, domain discriminators $\mathscr{D}$, batch indices $\mathcal{I}$
\State \textbf{Input}: $\mathbf{D}^s = [\mathbf{X}^{B^c};\mathbf{X}^{B};\mathbf{Y}^{B}]$, $\mathbf{X}^{A^c}$
\For{$e=1,2,...,E$}
\For{$i \in \mathcal{I}$}
    \State \textbf{Party} $\boldsymbol C$ do:
    \State \quad $\mathbf{x}^{A^c} \xleftarrow{}$ sample a mini-batch from $\mathbf{X}^{A^c}$;
    \State \quad $\mathbf{x}^{B^c} \xleftarrow{}$ select $i$th mini-batch from $\mathbf{X}^{B^c}$;
    \State \quad update models in $\mathscr{F}, \mathscr{D}$ by optimizing (\ref{argminmax_F_D}) \NoNumber{\quad using $\mathbf{x}^{A^c}$ and $\mathbf{x}^{B^c}$};
    \State \quad compute $\boldsymbol \mu^{B^c}$ by (\ref{mu_C}) using $\mathbf{x}^{B^c}$;
    \State \quad encrypt $\boldsymbol \mu^{B^c}$ and send $[[\boldsymbol\mu^{B^c}]]$ to party B;
    \State \textbf{Party} $\boldsymbol B$ do:
    \State \quad $\mathbf{x}^{B} \xleftarrow{}$ select $i$th min-batch from $\mathbf{X}^{B}$;
    \State \quad $\mathbf{y}^{B} \xleftarrow{}$ select $i$th min-batch from $\mathbf{Y}^{B}$;
    \State \textbf{Party} $\boldsymbol B$ and \textbf{Party} $\boldsymbol C$ do:
    % \State \quad SecureFederatedTraining($[[\boldsymbol\mu^{s,C}]], \mathbf{x}^{s,A}, \mathbf{y}^{s,A}$);
    \State \quad optimize (\ref{argmin_F_R}) using Algorithm \ref{sft} with \NoNumber{\quad $[[\boldsymbol\mu^{B^c}]], \mathbf{x}^{B}, \mathbf{y}^{B}$};
\EndFor
\EndFor

% $$$$
% \State \textbf{LocalDomainadaptation}($\mathbf{x}^{s,C}$, $\mathbf{x}^{t,C}$):
% \State \quad update models in $\mathscr{F}, \mathscr{D}$ by optimizing eq (\ref{argminmax_F_D});
% \State \quad $\boldsymbol \mu^{s,C} \xleftarrow{} g(\mathbf{x}^{s,C})$ by eq (\ref{mu_C}) and return $\boldsymbol \mu^{s,C}$;

% $$$$
% \State \textbf{SecureFederatedTraining}($[[\boldsymbol\mu^{s,C}]], \mathbf{x}^{s,A}, \mathbf{y}^{s,A}$):
% \State \quad perform SFP($[[\boldsymbol\mu^{s,C}]], \mathbf{x}^{s,A}, \mathbf{y}^{s,A}$) by Algorithm 2;
% \State \quad perform SBP() by Algorithm 3;

\end{algorithmic}
\end{algorithm}

\subsection{Fine-tuning Stage}

The fine-tuning stage aims to train the label predictor $R^A$ possessed by target party A using target labeled data $\mathbf{D}^{t}_l$. Prior to fine-tuning, party C initializes its feature extractors with pre-trained parameters. Note that since party A and party B are two independent parties, the trained label predictor $R^{B}$ of of source party B cannot be fine-tuned by $R^{A}$ of target party A. Thus, $R^{A}$ \textit{has to be trained from the scratch}. 

For each iteration, party C applies (\ref{mu_C}) to compute feature vectors $\boldsymbol\mu^{A^c}$ and then send $\boldsymbol\mu^{A^c}$ to party A for computing the label prediction loss:
\begin{multline}
\label{loss_bce_ft}
L_{ce}(\mathscr{F}, R^A) = \\ \mathbb{E}_{(\mathbf{x}^{A^c}_l,\mathbf{x}^{A}_l,\mathbf{y}^{A}) \sim \mathbf{D}^{t}_l} [\ell_{ce}(R^{A}([\boldsymbol\mu^{A^c};\mathbf{x}^{A}_l]), \mathbf{y}^{A})]
\end{multline}

% All model parameters are fine-tuned for minimizing  $L_{ce}(\mathscr{F}, R^B)$. 
Algorithm \ref{fed_fine_tune} describes the fine-tuning procedure and it is quite similar to Algorithm \ref{feddann_algo} except that it does not require party C to optimize (\ref{argminmax_F_D}).

\begin{algorithm}[h]
\caption{Federated fine-tuning}
\label{fed_fine_tune}
\begin{algorithmic}[1]
\State \textbf{Initialization}: feature extractors $\mathscr{F}$, batch indices $\mathcal{I}$
\State \textbf{Input}: $\mathbf{D}^t_l = [\mathbf{X}^{A^c}_l;\mathbf{X}^{A}_l;\mathbf{Y}^{A}]$
\For{$e=1,2,...,K$}
\For{$i \in \mathcal{I}$}
    \State \textbf{Party} $\boldsymbol C$ do:
    \State \quad $\mathbf{x}^{A^c} \xleftarrow{}$ select $i$th mini-batch from $\mathbf{X}^{A^c}_l$;
    \State \quad compute $\boldsymbol \mu^{A^c}$ by (\ref{mu_C}) using $\mathbf{x}^{A^c}$;
    \State \quad encrypt $\boldsymbol\mu^{A^c}$ and send $[[\boldsymbol\mu^{A^c}]]$ to party $\boldsymbol A$;
    \State \textbf{Party} $\boldsymbol A$ do:
    \State \quad $\mathbf{x}^{A}_l \xleftarrow{}$ select $i$th min-batch from $\mathbf{X}^{A}_l$;
    \State \quad $\mathbf{y}^{A} \xleftarrow{}$ select $i$th min-batch from $\mathbf{Y}^{A}$;
    \State \textbf{Party} $\boldsymbol A$ and \textbf{Party} $\boldsymbol C$ do:
    \State \quad minimize (\ref{loss_bce_ft}) using Algorithm \ref{sft} with \NoNumber{\quad $[[\boldsymbol\mu^{A^c}]], \mathbf{x}^{A}_l, \mathbf{y}^{A}$};
\EndFor
\EndFor
\end{algorithmic}
\end{algorithm}

\begin{algorithm}[h]
\caption{Privacy-preserving Federated Training}
\label{sft}
\begin{algorithmic}[1]
% \REQUIRE Model $\mathcal{M}_\mathcal{F}$, number of folds $K$
\State \textbf{Input}: $[[\boldsymbol\mu^{C}]], \mathbf{x}^{p}, \mathbf{y}^{p}$, where $p \in \{A, B\}$
% \State \textbf{SecureFederatedTraining}($[[\boldsymbol\mu^{s,C}]], \mathbf{x}^{s,A}, \mathbf{y}^{s,A}$):
\State \quad run Algorithm \ref{PPFP} with $[[\boldsymbol\mu^{C}]], \mathbf{x}^{p}, \mathbf{y}^{p}$;
\State \quad run Algorithm \ref{PPBP} with $[[\boldsymbol\mu^{C}]], \mathbf{x}^{p}$;

\end{algorithmic}
\end{algorithm}

\section{Privacy-preserving Vertical Federated Learning Framework}
\label{sfl}
As shown in (\ref{loss_bce}) (\ref{loss_bce_ft}), minimizing label prediction loss for training label predictor involves data from a active party (either party A or party B) and the passive party C. Therefore, the label predictor should be trained in a privacy-preserving manner. In this section, we elaborate our proposed privacy-preserving vertical federated learning framework (\textbf{PP-VFL}) of PrADA that enables two independent parties to collaboratively train the label predictor without exposing their private data. First, we define the label predictor model, which is LR, as follows: 
\begin{equation}
 \begin{aligned} 
% R^p([\mathbf{x}^{p};\boldsymbol\mu^{C}]) = \sigma([\boldsymbol\mu^{C}; \mathbf{x}^{p}]\mathbf{W}^p + b^p)
R^p([\boldsymbol\mu^{C};\mathbf{x}^{p}]) = \sigma([\boldsymbol\mu^{C}; \mathbf{x}^{p}]\mathbf{W} + b)
 \end{aligned} 
\end{equation}

where $\sigma$ is the sigmoid function and $p \in \{A, B\}$ denote an active party, $\mathbf{W} \in \mathbb{R}^{m+g} $ is the weights of model $R^p$ and $b \in \mathbb{R}^1$ is the bias. In this section, we denote $\boldsymbol\mu^{C}$ as the feature vectors from party C and $\mathbf{x}^{p}$ as raw features from the active party $p \in \{A,B\}$. We further decompose the input of $\sigma$ as follows:
% Since $[\boldsymbol\mu^{C};\mathbf{x}^{p}]$ is composed of data from the passive party C and a active party $p \in \{A, B\}$, we decompose the input of $\sigma$ as follows:
\begin{equation}
\label{logits}
z = \boldsymbol\mu^{C}\mathbf{W}^C +  \mathbf{x}^{p}\mathbf{W}^p + b^p
\end{equation}

where $\mathbf{W}^C \in \mathbb{R}^{g}$ is for the input $\boldsymbol\mu^{C}$ from party C while $\mathbf{W}^p \in \mathbb{R}^{m}$ is for the input $\mathbf{x}^{p}$ from party $p$. Both $\mathbf{W}^p$ and $\mathbf{W}^C$ are maintained by party $p$, but the real value of $\mathbf{W}^C$ is concealed from both party $p$ and party C, as elaborated in sections \ref{ppfp_sec} and \ref{ppbp_sec}.  

% so that party $p$ cannot infer private data of party C from $\mathbf{W}^C$. 

We extend the PHE-based secure protocol applied to the setting where one party has features and another has only labels \cite{Zhang2020AdditivelyHE} to our setting where features are distributed among two parties. Our new secure protocol includes two stages: (1) privacy-preserving forward propagation (Algo $\ref{PPFP}$) and (2) privacy-preserving backward propagation (Algo $\ref{PPBP}$). We denote the PHE encryption, addition and multiplication as $[[\cdot]]$, $\oplus$ and $\otimes$, respectively. Note that in our setting, only party C can encrypt and decrypt exchanging messages.

\subsection{Privacy-Preserving Forward Propagation}
\label{ppfp_sec}

Algorithm \ref{PPFP} aims to compute the label prediction loss in (\ref{loss_bce}) without compromising the data privacy of participating parties. To achieve this, the party C encrypts $\mu^C$ with PHE and sends encrypted $[[\mu^C]]$ to party $p$ to prevent privacy leakage. When receiving $[[\mu^C]]$, party $p$ can compute logit $z$ according to (\ref{logits}). However, directly applying (\ref{logits}) yields $[[z]]$ that is not compatible with logistic function. The workaround is that party $p$ first computes $[[\tilde{z}^C]]$ (Algo 4, line 5) and sends it to party C with random noise $\epsilon^p$ (Algo 4, line 8). The party C then decrypts $[[\tilde{z}^C + \epsilon^p]]$ and adds it with $\boldsymbol\mu^C \varepsilon^C_t$, which is for cancelling out the accumulated random noise, denoted as $\varepsilon^C_t$, that was embedded in $\widetilde{\mathbf{W}}_t^C$ during the backpropagation of the previous iteration. For now, we assume $\widetilde{\mathbf{W}}_t^C = \mathbf{W}_t^C - \varepsilon^C_t$, which we will prove in section \ref{ppbp_sec}. Here, we prove the logit $z^C$ is calculated correctly (Algo 4, line 11):
% \begin{equation}
% \begin{split}
% z &= z^A+ \tilde{z}^C + \boldsymbol\mu^C \varepsilon^C_t \\
%  &= \mathbf{x}^A \mathbf{W}_t^A + b + \boldsymbol\mu^C \widetilde{\mathbf{W}}_t^C + \boldsymbol\mu^C \varepsilon^C_t \\
%  &=\mathbf{x}^A \mathbf{W}_t^A + b + \boldsymbol\mu^C \mathbf{W}_t^C - \boldsymbol\mu^C \varepsilon^C_t + \boldsymbol\mu^C \varepsilon^C_t \\
%  &=\mathbf{x}^A \mathbf{W}_t^A + \boldsymbol\mu^C \mathbf{W}_t^C + b
% \end{split}
% \end{equation}
\begin{equation}
\label{z_C}
\begin{split}
z^C &= \tilde{z}^C + \boldsymbol\mu^C \varepsilon^C_t \\
 &=  \boldsymbol\mu^C \widetilde{\mathbf{W}}_t^C + \boldsymbol\mu^C \varepsilon^C_t \\
 &= \boldsymbol\mu^C \mathbf{W}_t^C - \boldsymbol\mu^C \varepsilon^C_t + \boldsymbol\mu^C \varepsilon^C_t \\
 &= \boldsymbol\mu^C \mathbf{W}_t^C 
\end{split}
\end{equation}
As a result, party C has $z^C + \epsilon^p$. The noise $\epsilon^p$ is for preventing party C from accessing the plaintext $z^C$ and further collecting $\mathbf{W}_t^C$ = $z^C / \boldsymbol\mu^C$. Finally, the party $p$ computes the loss $\ell_{ce}(\sigma(z), \mathbf{y}^{p})$.
\begin{algorithm}[ht]
\caption{Privacy-preserving Forward Propagation}
\label{PPFP}
\begin{algorithmic}[1]
\State \textbf{Initialization}: label predictor model $\widetilde{\mathbf{W}}_0^C$ and $\mathbf{W}_0^p$, accumulated noise $\varepsilon^C_0$
\State \textbf{Input}: $[[\boldsymbol\mu^{C}]], \mathbf{x}^{p}, \mathbf{y}^{p}$, $p \in \{A, B\}$
\State \textbf{Party} $\boldsymbol p$:
\State \quad compute logit:
\State \quad\quad\quad\quad $[[\tilde{z}^C]] \xleftarrow{} [[\boldsymbol\mu^C]] \otimes \widetilde{\mathbf{W}}_t^C$;
\State \quad\quad\quad\quad $z^p \xleftarrow{} \mathbf{x}^p \mathbf{W}_t^p + b^p$;
% \State \quad generate random noise $\epsilon^A$;
\State \quad add noise $[[\tilde{z}^C + \epsilon^p]] \xleftarrow{} [[\tilde{z}^C]] \oplus \epsilon^p$;
\State \quad send $[[\tilde{z}^C + \epsilon^p]]$ to party $\boldsymbol C$;
\State \textbf{Party} $\boldsymbol C$:
\State \quad $\tilde{z}^C + \epsilon^p \xleftarrow{}$ decrypt $[[\tilde{z}^C + \epsilon^p]]$;
% \State \quad add $\boldsymbol\mu^C \varepsilon^C_t$ to $\tilde{z}^C + \epsilon^A$;
\State \quad $z^C + \epsilon^p \xleftarrow{}$ $\tilde{z}^C$ + $\boldsymbol\mu^C \varepsilon^C_t$ + $\epsilon^p$;
\State \quad send $z^C + \epsilon^p$ to party $\boldsymbol p$;
\State \textbf{Party} $\boldsymbol p$:
\State \quad remove noise $z^C \xleftarrow{} z^C + \epsilon^p$;
\State \quad $z \xleftarrow{}$ $z^p+ z^C$;
\State \quad compute loss $\ell_{ce}(\sigma(z), \mathbf{y}^{p})$;
\end{algorithmic}
\end{algorithm}

\begin{algorithm}[ht]
\caption{Privacy-preserving Backward Propagation}
\label{PPBP}
\begin{algorithmic}[1]
% \Require Model parameters $\Theta^A$, $\Theta^B$, $\{x_j^B\}_{j \in N_B}$
\State \textbf{Initialization}: learning rate $\eta$
\State \textbf{Input}: $[[\boldsymbol\mu^{C}]], \mathbf{x}^{p}$, $p \in \{A, B\}$
\State \textbf{Party} $\boldsymbol p$:
\State \quad $\delta^{l} \xleftarrow{}$ $\nabla_{\sigma}\ell_{ce}$ w.r.t the activation function $\sigma$;
\State \quad backpropagate gradients $\delta^{l}$:
\State \quad\quad\quad\quad $[[\Delta\mathbf{W}^C_t]] \xleftarrow{} [[\boldsymbol\mu^C]] \otimes \delta^{l}$;
\State \quad\quad\quad\quad $\Delta\mathbf{W}^p_t \xleftarrow{} \mathbf{x}^{p} \delta^{l}$;
\State \quad\quad\quad\quad $\Delta b^p_t \xleftarrow{} \delta^{l}$;
% \State \quad generate random noise $\epsilon^A$;
\State \quad add noise $[[\Delta\mathbf{W}^C_t + \epsilon^p]] \xleftarrow{} [[\Delta\mathbf{W}^C_t]] \oplus \epsilon^p$;
\State \quad send $[[\Delta\mathbf{W}^C_t + \epsilon^p]]$ to party $\boldsymbol C$;
\State \textbf{Party} $\boldsymbol C$:
\State \quad $\Delta\mathbf{W}^C_t + \epsilon^p \xleftarrow{}$ decrypt $[[\Delta\mathbf{W}^C_t + \epsilon^p]]$;
\State \quad add noise $\Delta\widetilde{\mathbf{W}}^C_t + \epsilon^p \xleftarrow{}$ $\Delta\mathbf{W}^C_t + \frac{\epsilon^C}{\eta} + \epsilon^p$;
\State \quad $\varepsilon^C_{t+1} \xleftarrow{} \varepsilon^C_{t} + \epsilon^C$ and $[[\varepsilon^C_{t+1}]] \xleftarrow{}$ encrypt $\varepsilon^C_{t+1}$;
\State \quad send $[[\varepsilon^C_{t+1}]]$, $\Delta\widetilde{\mathbf{W}}^C_t + \epsilon^p$ to party $\boldsymbol p$ 
\State \textbf{Party} $\boldsymbol p$:
\State \quad remove noise $\Delta\widetilde{\mathbf{W}}^C_t \xleftarrow{} \Delta\widetilde{\mathbf{W}}^C_t + \epsilon^p$
\State \quad update weights and bias of logistic regression model:
\State \quad\quad\quad\quad $\widetilde{\mathbf{W}}^C_{t+1} \xleftarrow{} \widetilde{\mathbf{W}}^C_{t} - \eta \Delta\widetilde{\mathbf{W}}^C_{t}$;
\State \quad\quad\quad\quad $\mathbf{W}^p_{t+1} \xleftarrow{} \mathbf{W}^p_{t} - \eta \Delta\mathbf{W}^p_{t}$;
\State \quad\quad\quad\quad $b_{t+1}^p \xleftarrow{} b_{t}^p - \eta \Delta b_{t}^p$;
\State \quad $[[\delta^C]] \xleftarrow{} \delta^l \otimes (\widetilde{\mathbf{W}}^C_{t+1} \oplus [[\varepsilon^C_{t+1}]])$;
\State \quad send $[[\delta^C]]$ to party $\boldsymbol C$;
\State \textbf{Party} $\boldsymbol C$:
\State \quad $\delta^C \xleftarrow{}$ decrypt $[[\delta^C]]$;
\State \quad update feature aggregators in $\mathscr{G}$ and feature extractors 
\State \quad in $\mathscr{F}$ based on gradient $\delta^C$ using SGD;
\end{algorithmic}
\end{algorithm}

\subsection{Privacy-Preserving Backward Propagation}
\label{ppbp_sec}
During the privacy-preserving backward propagation as described in Algorithm \ref{PPBP}, the active party $p$ securely updates logistic regression model $R^p$ and backpropagate gradients to party C. As shown in (\ref{logits}), we partitioned weights of $R^p$ into $\mathbf{W}^p$ and $\mathbf{W}^C$. On the one hand, party $p$ can compute gradients $\Delta \mathbf{W}^p_t$ and $\Delta b^p_t$, and update $\mathbf{W}^p_t$ and $b^p_t$ (Algo 5, line 20-21) in plaintext since party $p$ owns these parameters. On the other hand, party A can not directly update $[[\mathbf{W}^C_{t+1}]] \xleftarrow{} \mathbf{W}^C_t -  \eta [[\Delta\mathbf{W}^C_t]]$ since this leads to incompatibility with PHE for computing $[[\tilde{z}^C]] \xleftarrow{} [[\boldsymbol\mu^C]] \otimes [[\mathbf{W}^C_{t+1}]]$ in the next iteration of forward propagation. To work around this issue, party $p$ may send encrypted gradients $[[\Delta\mathbf{W}^C_{t}]]$ to party C and get the decrypted $\Delta\mathbf{W}^C_{t}$ back. However, this leads to privacy leakage for both parties. Because based on $\Delta\mathbf{W}^C_t = \boldsymbol\mu^C \otimes \delta_{l}$, knowing $\Delta\mathbf{W}^C_{t}$ party $p$ can infer the value of $\boldsymbol\mu^C$, while party C can infer gradient $\delta_l$ during training. 
Therefore, to conceal the real value of $\Delta\mathbf{W}^C_{t}$ from both parties, the two parties mask $\Delta\mathbf{W}^C_{t}$ by adding corresponding random noises. Specifically, party $p$ adds noise $\epsilon^p$ to $[[\Delta\mathbf{W}^C_{t}]]$ and sends $[[\Delta\mathbf{W}^C_{t}+\epsilon^p]]$ to party C (Algo 5, line 9-10). Party C in turn decrypts $[[\Delta\mathbf{W}^C_{t}+\epsilon^p]]$ and sends $\Delta\widetilde{\mathbf{W}}^C_t + \epsilon^p$ back to party $p$, where $\Delta\widetilde{\mathbf{W}}^C_t = \Delta\mathbf{W}^C_t + \frac{\epsilon^C}{\eta} $ (Algo 5, line 13), $\epsilon^C$ is the random noise generated by party C and $\eta$ is the learning rate. Then, the party $p$ updates $\widetilde{\mathbf{W}}^C_{t+1}$ based on gradient $\Delta\widetilde{\mathbf{W}}^C_t$ after removing noise $\epsilon^p$  (Algo 5, line 19). Note that while the noise $\epsilon^p$ can be removed by party $p$, the noise $\epsilon^C$ added by party C is accumulated in weight $\widetilde{\mathbf{W}}^C_t$ through $\Delta\widetilde{\mathbf{W}}^C_t $ at each iteration. Intuitively, the real value of $\mathbf{W}^C_{t+1}$ can be seem as shared by party C and party $p$. This concept is similar to secret sharing.

For party $p$ to correctly calculate the intermediate gradient $\delta^C$, party $p$ needs to cancel out the accumulated noise embedded in $\widetilde{\mathbf{W}}^C_{t+1}$. To this end, party C seeds the encrypted accumulated noise $[[\varepsilon^C_{t+1}]]$ to party $p$, which then calculates the gradient $[[\delta^C]]$ of loss $\ell_{ce}$ with respect to $\mu^C$ (Algo 5, line 22) using $[[\varepsilon^C_{t+1}]]$.  
% For party $p$ to cancel out the accumulated noise embedded in $\widetilde{\mathbf{W}}^C_{t+1}$, party C needs to seed the encrypted accumulated noise $[[\varepsilon^C_{t+1}]]$ to party $p$. Then, party A calculates the gradient $[[\delta^C]]$ of loss $\ell_{ce}$ with respect to $\mu^C$ (Algo 5, line 22) using $[[\varepsilon^C_{t+1}]]$. 
To prove the value of gradient $\delta^C$ is calculated correctly, we prove that $\widetilde{\mathbf{W}}^C_{t+1}=\mathbf{W}^C_{t+1}-\varepsilon^C_{t+1}$ by mathematical induction, assuming $\widetilde{\mathbf{W}}^C_{t}=\mathbf{W}^C_{t} - \varepsilon^C_{t}$ and Initializing $\varepsilon^C_{0}=0$:
\begin{equation*}
\begin{split}
    \widetilde{\mathbf{W}}^C_{t+1} &= \widetilde{\mathbf{W}}^C_{t} - \eta \Delta\widetilde{\mathbf{W}}^C_{t} \\
    &=\widetilde{\mathbf{W}}^C_{t} - \eta( \Delta\mathbf{W}^C_{t}+\frac{\epsilon^C}{\eta} ) \\
    &=(\mathbf{W}^C_{t}-\eta\Delta\mathbf{W}^C_{t}) - (\varepsilon^C_{t} + \epsilon^C) \\
    &=\mathbf{W}^C_{t+1}-\varepsilon^C_{t+1}
\end{split}
\end{equation*}

Finally, party $p$ sends $[[\delta^C]]$ back to party C, which decrypts $[[\delta^C]]$ and backpropagates $\delta^C$ locally to optimize local models.

\begin{table*}[tb]
\caption{Comparison between models in different settings on Census Income dataset.}
\label{exp_census}
\centering
\begin{tabular}{l|l||c|c||c|c||c|c}
\hline
\multicolumn{2}{c}{Positive labels} & \multicolumn{2}{c}{40} & \multicolumn{2}{c}{80} & \multicolumn{2}{c}{160} \\
\hline
\multicolumn{1}{c}{Setting} & \multicolumn{1}{l}{Model} & \multicolumn{1}{c}{AUC (\%)} & \multicolumn{1}{c}{KS (\%)} & \multicolumn{1}{c}{AUC (\%)} & \multicolumn{1}{c}{KS (\%)} & \multicolumn{1}{c}{AUC (\%)} & \multicolumn{1}{c}{KS (\%)} \\
\hline
\hline
 $\mathbf{A}$-$\mathbf{Local}$ & LR & 65.57 & 31.48 & 72.33 & 35.32 & 72.71 & 35.45\\
    & XGBoost & 67.59 & 32.27 & 74.07 & 37.62 & 77.60 & 41.68 \\
\hline
 $\mathbf{A}$-$\mathbf{VFL}$ & SecureLR & 69.67 & 32.21 & 72.72 & 36.87  & 75.07 & 39.48 \\
     & SecureBoost & 71.97 & 34.73 & 77.02 & 41.61  & 80.08 & 46.80 \\
     & $\text{PrADA}_{\text{w/o DA\&FG\&IR}}$ & 73.72$\pm$0.41 & 35.36$\pm$0.68 & 77.48$\pm$0.47 & 42.32$\pm$0.30 & 79.13$\pm$0.68 & 44.90$\pm$0.57 \\
\hline
$\mathbf{AB}$-$\mathbf{VFL}$ & SecureLR & 72.88 & 34.73 & 73.80 & 35.83 & 74.63 & 38.48\\
      & SecureBoost & 78.06 & 42.18 & 79.56 & 45.56 & 80.82 & 47.87\\
     & $\text{PrADA}_{\text{w/o DA\&FG\&IR}}$ & 77.65$\pm$0.38 & 43.09$\pm$0.64 & 78.97$\pm$0.49 & 45.91$\pm$0.51 & 80.56$\pm$0.31 & 47.62$\pm$0.48 \\
\hline
\hline
% $\checkmark$ & $\checkmark$ & $\text{PrADA}_{\text{w/o FG}}$ & 80.01 & 46.65 \\
 $\mathbf{B} \rightarrow \mathbf{A}$ & $\text{PrADA}_{\text{w/o FG\&IR}}$ & 78.98$\pm$0.13 & 43.42$\pm$0.50 & 80.17$\pm$0.28 & 46.86$\pm$0.93 & 81.10$\pm$0.57 & 48.14$\pm$0.75 \\
    & $\text{PrADA}_{\text{w/o IR}}$ & 78.92$\pm$0.16 & 44.06$\pm$0.72  & 80.49$\pm$0.37 & 47.36$\pm$0.55 & 81.36$\pm$0.15 & 48.73$\pm$0.56   \\
    & $\text{PrADA}$ & \textbf{79.17$\pm$0.40} & \textbf{44.92$\pm$0.68}  & \textbf{81.08$\pm$0.30} & \textbf{48.06$\pm$0.72} & \textbf{81.46$\pm$0.06} & \textbf{49.27$\pm$0.42}   \\
\hline
\end{tabular}
\end{table*}

\subsection{Discussions on Privacy Protection}
% In this section, we discuss the privacy-preserving capability of our privacy-preserving vertical federated learning framework (PP-VFL) and its possible privacy leakage and trade-off. 

In this section, we discuss the privacy-preserving capability of our PP-VFL, its possible privacy leakage and trade-off. 

\begin{prop}
\label{prop_1}
The active party $p$ cannot reveal the true value of the feature vectors $\boldsymbol\mu^C$ passed from the passive party C during training and inference.
\end{prop}

\emph{Proof.}{
There are three ways through which party $p$ can leverage to recover the true value feature vectors $\boldsymbol\mu^C$ during training. The first way is to decrypt $[[\boldsymbol\mu^C]]$ directly. However, it is impossible to decrypt $[[\boldsymbol\mu^C]]$ without knowing the private key. The second way is to derive $\boldsymbol\mu^C$ from $z^C/\widetilde{\mathbf{W}}_t^C$ according to (\ref{z_C}). However, this requires party $p$ to remove the noise $\varepsilon^C_{t}$ from $\widetilde{\mathbf{W}}_t^C$. Suppose $\varepsilon^C_t$ is the random noise accumulated by party C at iteration $t$ and $\hat{\varepsilon}^C_t$ is an attempt from party A. The probability that $\varepsilon^C_t=\hat{\varepsilon}^C_t$ is $Pr(\varepsilon^C_t=\hat{\varepsilon}^C_t) \leq (1 - e^{-2/ \mathbb{|Z|}})$~\cite{Bruce2007}. Because $\mathbb{|Z|}$ is typically a very large number, the $Pr(\varepsilon^C_t=\hat{\varepsilon}^C_t)$ is very close to zero. Third, $\boldsymbol\mu^C$ can also be derived from $\Delta\mathbf{W}^C_t / \delta^{l}$ after the noise $\epsilon^C$ being removed from $\Delta\widetilde{\mathbf{W}}^C_t$. However, the probability $Pr(\epsilon^C=\hat{\epsilon}^C)$ that the attempt $\hat{\epsilon}^C$ made at party $p$ equals $\epsilon^C$ approximates zero.}

\begin{prop}
\label{prop_2}
The active party $p$ cannot infer the true value of weights $\mathbf{W}^C$ during training and inference.
\end{prop}

\emph{Proof.}{
% There two ways that party $p$ can infer $\mathbf{W}^C$. One is via $z^C /  \boldsymbol\mu^C$ according to (\ref{z_C}), and another is removing noise from $\widetilde{\mathbf{W}}^C$. The only information that party $p$ receives from party C is the feature vectors $\boldsymbol\mu^C$, and according to Proposition \ref{prop_1} the true value of $\boldsymbol\mu^C$ is concealed from the party $p$ during training and inference. Therefore, party $p$ infer the true value of $\mathbf{W}^C$  during training and inference.
There two ways that party $p$ can infer $\mathbf{W}^C$. One is via $z^C /  \boldsymbol\mu^C$ according to (\ref{z_C}), and another is removing accumulated noise $\varepsilon^C$ from $\widetilde{\mathbf{W}}^C$. According to Proposition \ref{prop_1} the true value of $\boldsymbol\mu^C$ is concealed from the party $p$ during training and inference, and the probability $Pr(\varepsilon^C=\hat{\varepsilon}^C) \leq (1 - e^{-2/ \mathbb{|Z|}})$ that the party C can generate noise $\hat{\varepsilon}^C$ to cancel out $\varepsilon^C$ is close to zero. Therefore, the party $p$ cannot infer the true value of $\mathbf{W}^C$  during training and inference.
}

The active party $p$ cannot infer the data of passive party C during training because party $p$ has no access to $\boldsymbol\mu^C$, $\mathbf{W}^C$ and party C's local model. \cite{he@2019mi} proposes model inversion (MI) that enables the attacker (i.e., party $p$) to recover the private data of the victim (i.e., party C) during inference. To recover data of reasonably high quality, \cite{he@2019mi} makes strong assumptions that the attacker knows the network structure of the victim and has access to the training data that follows the same distribution as those of the victim, aiming to approximate the victim's local model. However, these assumptions typically do not hold in scenarios like finance. For one thing, financial data generally are not publicly available because they are sensitive, and thus their publication is regulated. For another, participating parties provide heterogeneous features in VFL, and thereby they typically adopt different model structures. Besides,~\cite{he@2019mi} demonstrates that a local model with a full-connected layer on top can significantly drop the quality of the recovered data. The $\mathbf{W}^C$ automatically provides such a layer of protection. 
% Therefore, we should take the cost of the attacker's obtaining these data into account when we measure ... .
\begin{prop}
\label{prop_3}
The passive party C cannot infer the true value of weights $\mathbf{W}$ curated by the active party $p$ during training and inference.
\end{prop}

\emph{Proof.}{
The weights $\mathbf{W}$ is composed of $\mathbf{W^C}$ and $\mathbf{W}^p$. Party C receives no information on $\mathbf{W}^p$ of party $p$. Therefore, party C can learn nothing on $\mathbf{W}^p$. There are two ways that party C can infer $\mathbf{W}^C$. The first one is via $(z^C+\epsilon^p)/  \boldsymbol\mu^C$ and the another is via $\delta^C/ \delta^l$. The noise $\epsilon^p$ in the former one prevents party C from revealing $\mathbf{W^C}$ because the probability $Pr(\epsilon^p=\hat{\epsilon}^p) \leq (1 - e^{-2/ \mathbb{|Z|}})$ that the party C can generate noise $\hat{\epsilon}^p$ to cancel out $\epsilon^p$ is close to zero, while $\delta^l$ in the latter one resides only in party $p$. Therefore, the party C cannot reveal the true value of $\mathbf{W^C}$.
}

% \begin{prop}
% \label{prop_4}
% The passive party C cannot infer the data and labels of party $p$ during training and inference.
% \end{prop}

% \emph{Proof.}{
% Party C receives no information on the data $\mathbf{X}^p$ of active party $p$. Therefore, party C can learn nothing on $\mathbf{X}^p$. The only information on the labels of party $p$ that party C receives is gradient $\delta^{C}$, which can be used to infer gradient $\delta^{l}$ = $\delta^{C}/\mathbf{W^C}$. However, according to Proposition \ref{prop_3}, party C cannot infer  $\mathbf{W^C}$ and thus party C can learn nothing about the labels of party $p$.  During inference, party C gets nothing from party $p$. Therefore, party C cannot reveal data and labels of party $p$ during both training and inference.
% }

Recent research works propose that the attacker (i.e., passive party C) can leverage gradient inversion (GI)~\cite{zhu@2019dlg}, model completion (MC)~\cite{fu@2022mc}, and properties of cut-layer gradient (PCG) \cite{li2022label} to recover labels of the victim (i.e., active party $p$). Our PP-VFL can prevent GI attack because the attacker has no access to both the weights (i.e., $\mathbf{W}$) and the gradient of the label predictor model~\cite{kang@2022fedcg,Gu@2021splitfed}. While PP-VFL itself cannot prevent the MC and PCG attacks in that the cut-layer gradient $\delta^C$ is passed to the attacker in plain text and without any protection. The current form of PP-VFL trades a certain degree of increased label privacy leakage with enhanced model performance and training efficiency. In applications where the labels are important assets, the PP-VFL can be equipped with other privacy protection mechanisms (e.g., MARVELL~\cite{li2022label} for PCG and CoAE~\cite{Liu@2021coae} for MC) to trade the protection of label privacy with a certain degree of degraded utility.
% According to the no-free lunch for the privacy-utility trade-off~\cite{}
% \emph{Proof.}{
% Party C receives no information on the data $\mathbf{X}^p$ of active party $p$. Therefore, party C can learn nothing on $\mathbf{X}^p$. The only information on the labels of party $p$ that party C receives is the gradient $\delta^{l}$, which can be derived from $\Delta\mathbf{W}^C_t / \boldsymbol\mu^C$ (Algorithm \ref{PPBP}). However, party C only knows $\Delta\mathbf{W}^C_t + \epsilon^p$ and the probability $Pr(\epsilon^p=\hat{\epsilon}^p) \leq (1 - e^{-2/ \mathbb{|Z|}})~$\cite{Bruce2007} that party C can generate noise $\hat{\epsilon}^p$ that cancels out $\epsilon^p$ is close to zero.
% During inference, party C gets nothing from party $p$. Therefore, party C cannot reveal data and labels of party $p$.
% }

% Note that because we require the label predictor $R^B$ to be interpretable to party B,  party C needs to send the accumulated random noise to party B for recovering the real value of $W^C$ after fine-tuning procedure. When making inference, party B may either adopt Fully Homomorphic Encryption or Taylor series approximation to make the logit [[$z$]] compatible with sigmoid function. In applications where both real-time inference and accuracy are critical, $z$ can be calculated in plaintext at the cost of possibly information leakage.

\section{Experiments}

\subsection{Experimental datasets and settings}

We evaluate our proposed $\text{PrADA}$ based on two datasets: one is Census Income dataset, and another is a real-world financial dataset called Loan Default. For each dataset, we run experiments under following four settings:
\begin{enumerate}
    \item $\mathbf{A}$-$\mathbf{Local}$: Target party A only uses its local data to train models without leveraging VFL and DA.
    \item $\mathbf{A}$-$\mathbf{VFL}$: Target Party A uses target domain data $\mathbf{D}^t_l$ to train models via VFL with party C. This setting serves as the conventional VFL to improve the model performance of party A with additional features from party C. 
    % $\mathbf{D}^t = [\mathbf{X}^{A^c};\mathbf{X}^{A};\mathbf{Y}^{A}]$
    % This serves as the conventional VFL to improve the model performance of party A with additional features from party C. 
    % We train $\text{PrADA}_{\text{w/o DA}}$ (no domain discriminators applied) to serve as a lower bound for PrADA.
    \item $\mathbf{AB}$-$\mathbf{VFL}$: Assuming party A and B are from the same organization and privacy is not a concern. Thus, Party A uses $\mathbf{D}^t_l$ and $\mathbf{D}^s$ of both domains to train models via VFL with party C (with no DA). Models in this setting serve as strong baselines because they use all data together.
    % This scheme is to verify whether the performance of target model can be improved with additional samples. We train $\text{PrADA}_{\text{w/o DA}}$ to serve as a strong baseline (it uses all data) to evaluate the effectiveness of the DA approach applied in PrADA.
    \item $\mathbf{B} \rightarrow \mathbf{A}$: We conduct PrADA elaborated in sections \ref{fada} to perform federated adversarial domain adaptation from party B to party A.
    % and evaluate the performance of PrADA and its variants.
    % and compare PrADA with $\text{PrADA}_{\text{w/o FG}}$ (no feature grouping applied) to evaluate the effectiveness of FG on domain adversarial training. To fairly verify the effectiveness of our DA applied in $\text{PrADA}$, we evaluate a PrADA variant called $\text{PrADA}_{ft}$ that fine-tunes the target label predictor initialized with parameters of the pre-trained source label predictor.
\end{enumerate}

In settings 2 and 3, we adopt SecureLR and SecureBoost implemented in FATE\footnote{https://github.com/FederatedAI/FATE}, an industrial grade federated learning framework, as comparing models. These two models are VFL version of tree-boosting model and logistic regression model respectively, and they are using PHE to protect data privacy. To explore the effectiveness of different components of PrADA, we propose three different ablations:
% $\text{PrADA}_{\text{w/o DA\&FG\&IR}}$ without domain adaptation, feature grouping and interaction; $\text{PrADA}_{\text{w/o FG\&IR}}$ applies domain adaptation, but without feature grouping and interaction; $\text{PrADA}_{\text{w/o IR}}$ applies domain adaptation and feature grouping, but without interaction. 

\begin{itemize}
    \item  $\text{PrADA}_{\text{w/o DA\&FG\&IR}}$ without domain adaptation (DA), feature grouping (FG) and feature group interaction (IR);
    \item $\text{PrADA}_{\text{w/o FG\&IR}}$ applies domain adaptation, but without feature grouping and interaction;
    \item $\text{PrADA}_{\text{w/o IR}}$ applies domain adaptation based on feature grouping, but without interaction.
\end{itemize}

In this paper, we focus on the binary classification problem. Because imbalanced class label is one of the major motivations for applying domain adaptation in real-world financial applications, we also investigate the effectiveness of our PrADA approach for different positive label ratios. Specifically, we investigate scenarios when the target training data has a positive label ratio \{0.01,0.02,0.04\}. 

For SecureLR, we use default hyperparameters, while for SecureBoost, we sweep over all combination of \textit{max depth} \{2,4,6,8\} and \textit{number of trees} \{100,200,300,400\}, leaving other hyperparameters default. For PrADA, we use \textit{batch size} 128 for Census Income data and 64 for Loan Default data, and use \textit{learning rate} 0.0005 for pre-training and 0.0008 for fine-tuning for both datasets. Our PrADA is implemented with PyTorch. We repeat every experiment 5 times on each dataset, reporting the mean and standard derivation of AUC and KS (Kolmogorov-Smirnov test)~\cite{kst} of all trained models on \textit{test data of the target party A}.

\begin{figure*}[ht]
\centering
\subfigure[Non-adapted employment]{
\begin{minipage}[t]{0.24\linewidth}
\centering
\includegraphics[width=1.8in]{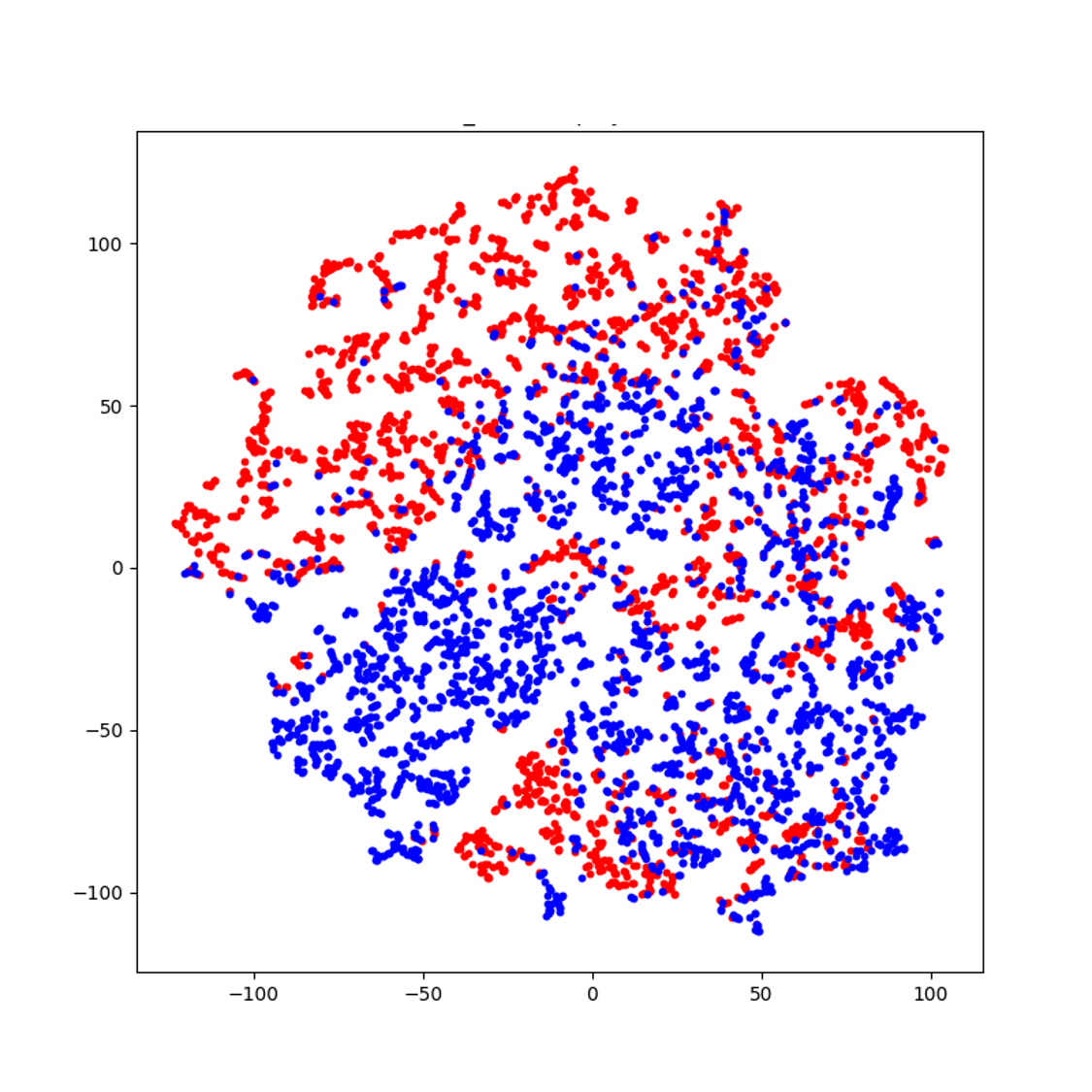}
%\caption{no_dann_employment}
\end{minipage}%
}%
\subfigure[Non-adapted demographics]{
\begin{minipage}[t]{0.24\linewidth}
\centering
\includegraphics[width=1.8in]{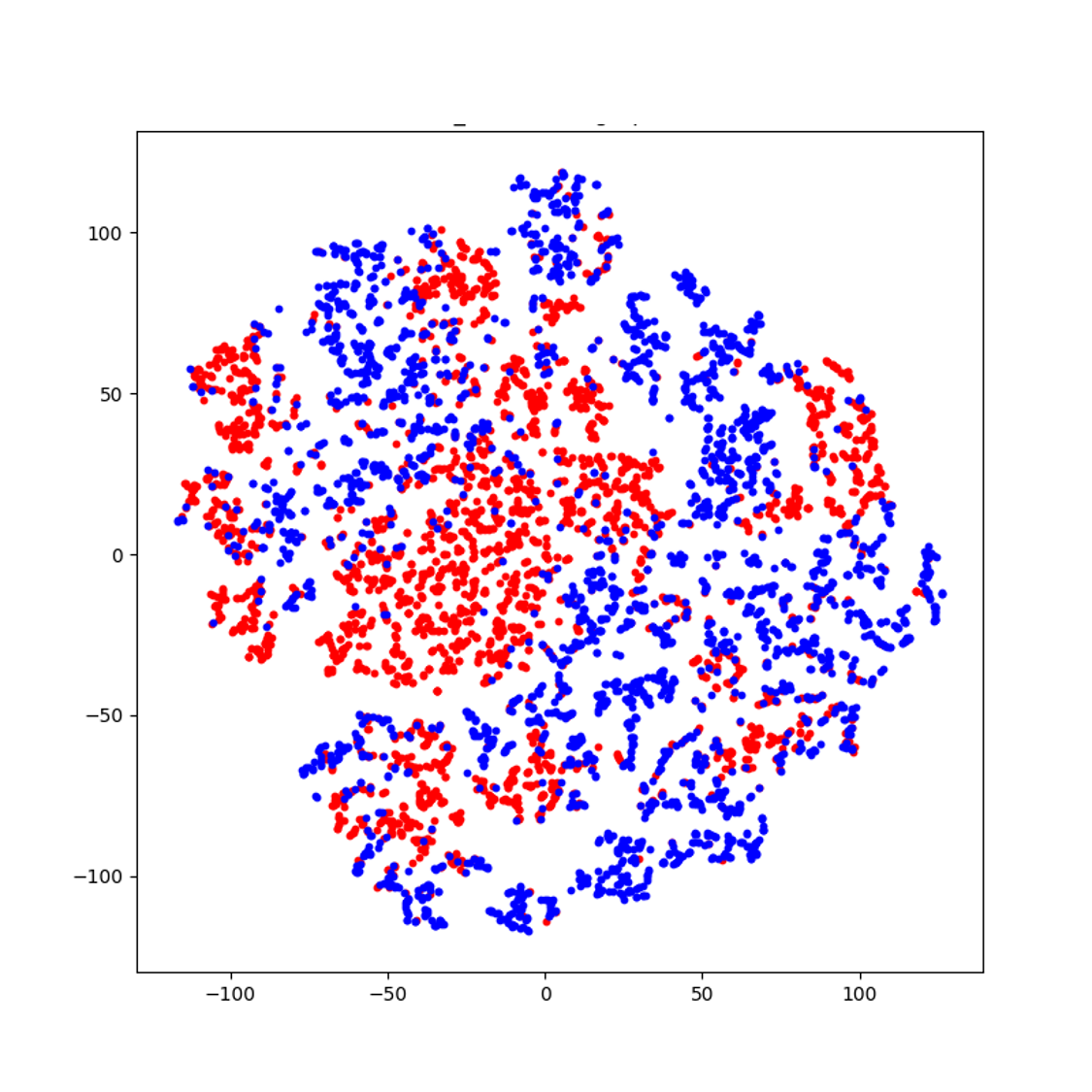}
%\caption{dann_employment}
\end{minipage}%
}%
\subfigure[Non-adapted household]{
\begin{minipage}[t]{0.24\linewidth}
\centering
\includegraphics[width=1.8in]{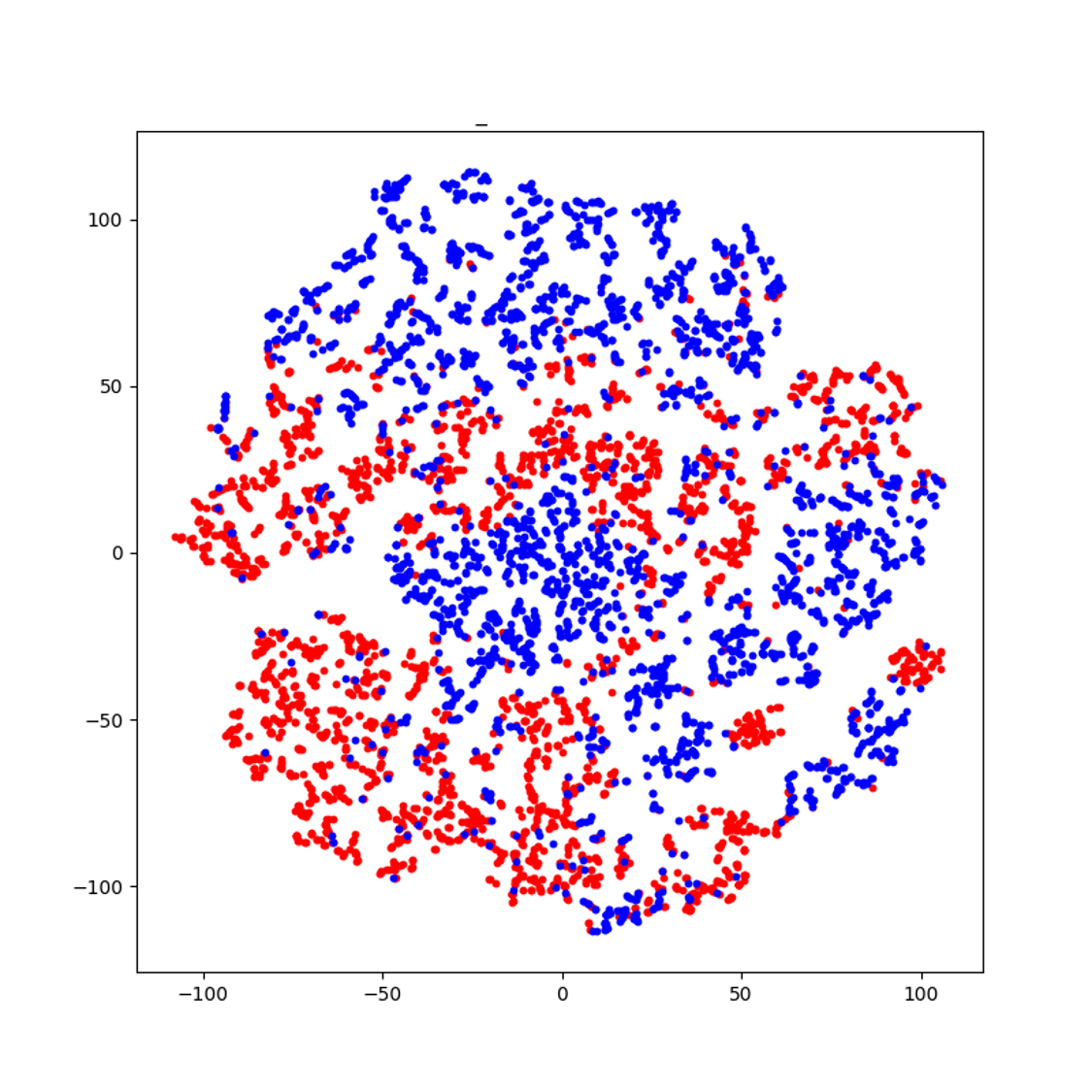}
%\caption{no_dann_demographics}
\end{minipage}
}%
\subfigure[Non-adapted migration]{
\begin{minipage}[t]{0.24\linewidth}
\centering
\includegraphics[width=1.8in]{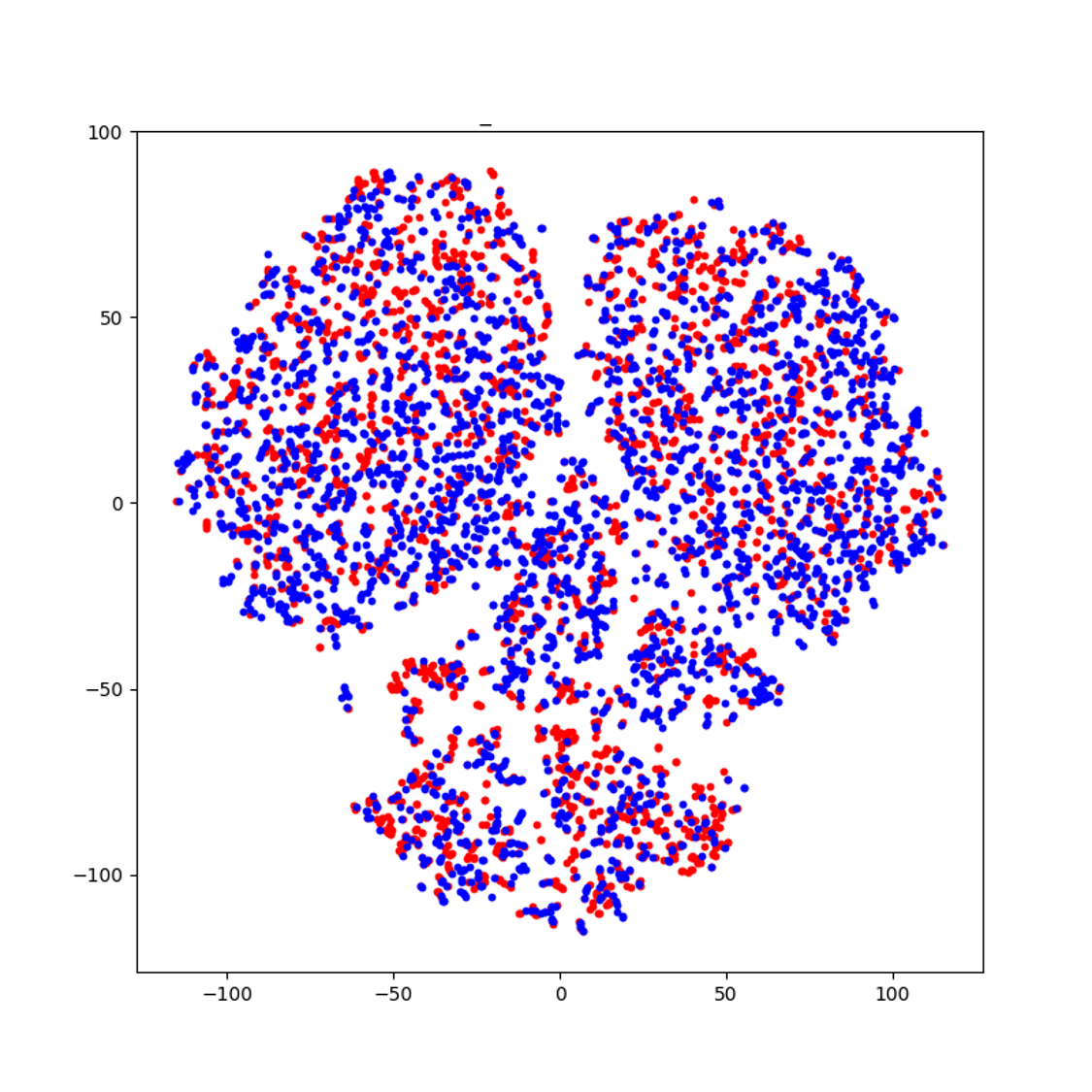}
%\caption{dann_demographics}
\end{minipage}
}%

\centering
\subfigure[Adapted employment]{
\begin{minipage}[t]{0.24\linewidth}
\centering
\includegraphics[width=1.8in]{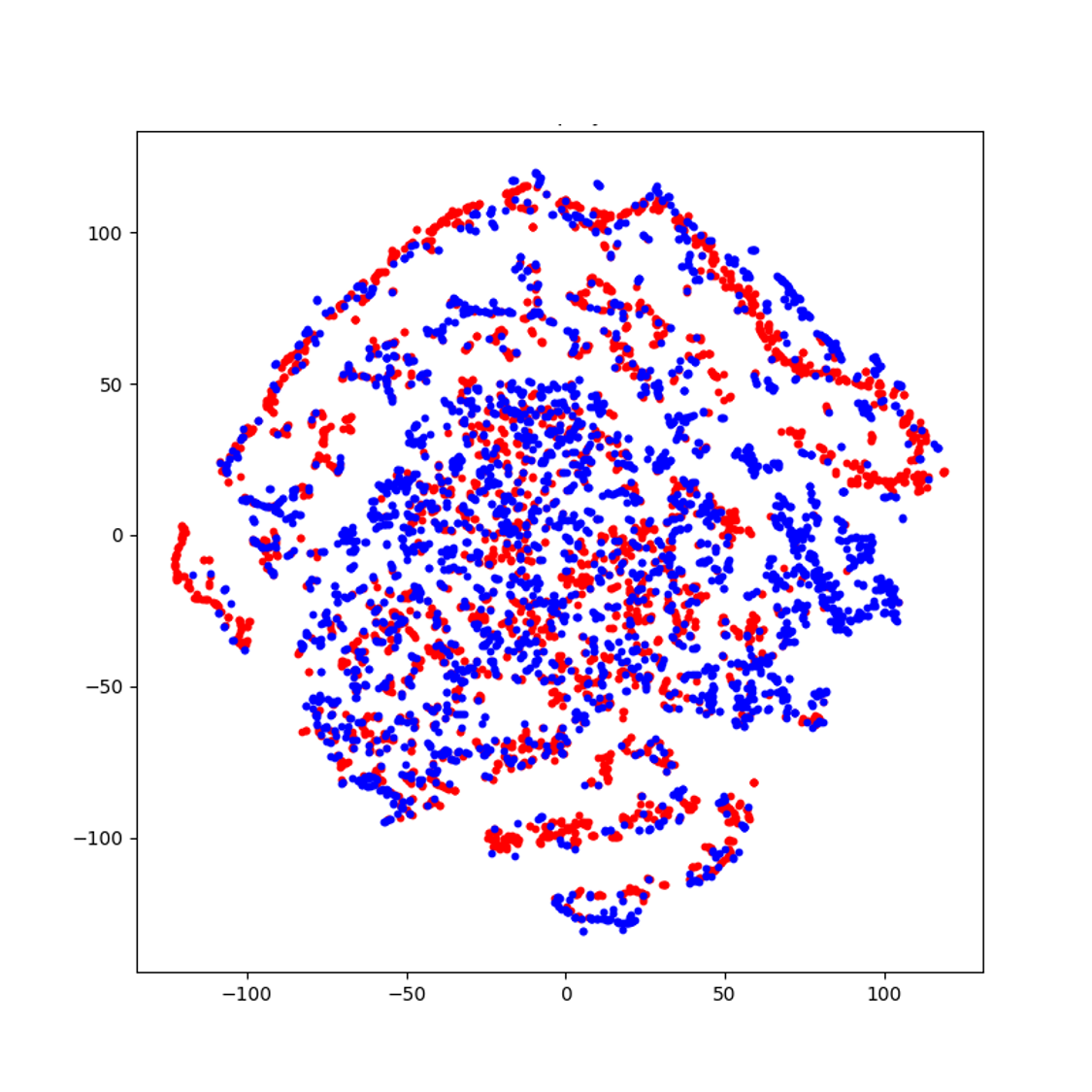}
%\caption{no_dann_employment}
\end{minipage}%
}%
\subfigure[Adapted demographics]{
\begin{minipage}[t]{0.24\linewidth}
\centering
\includegraphics[width=1.8in]{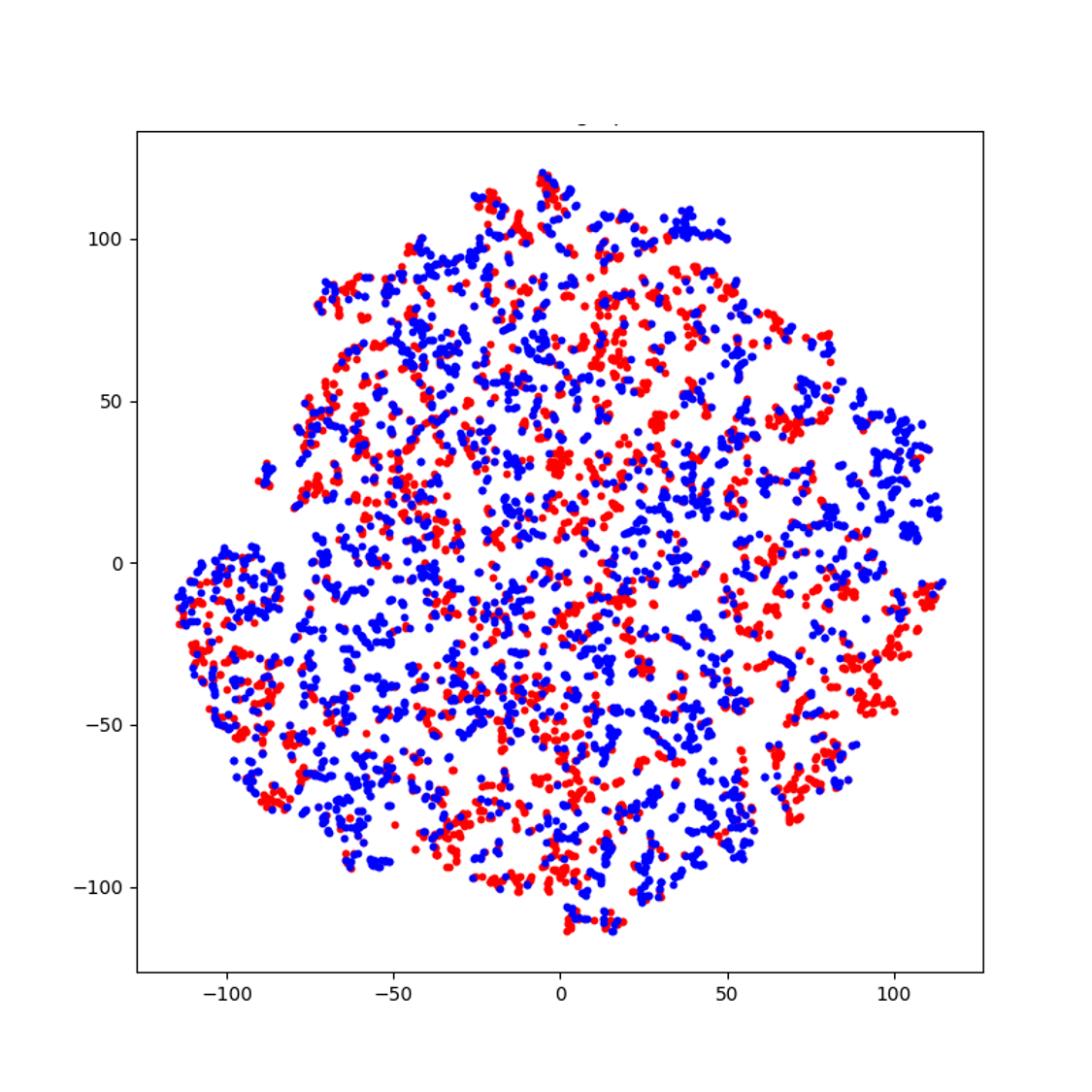}
%\caption{dann_employment}
\end{minipage}%
}%
\subfigure[Adapted household]{
\begin{minipage}[t]{0.24\linewidth}
\centering
\includegraphics[width=1.8in]{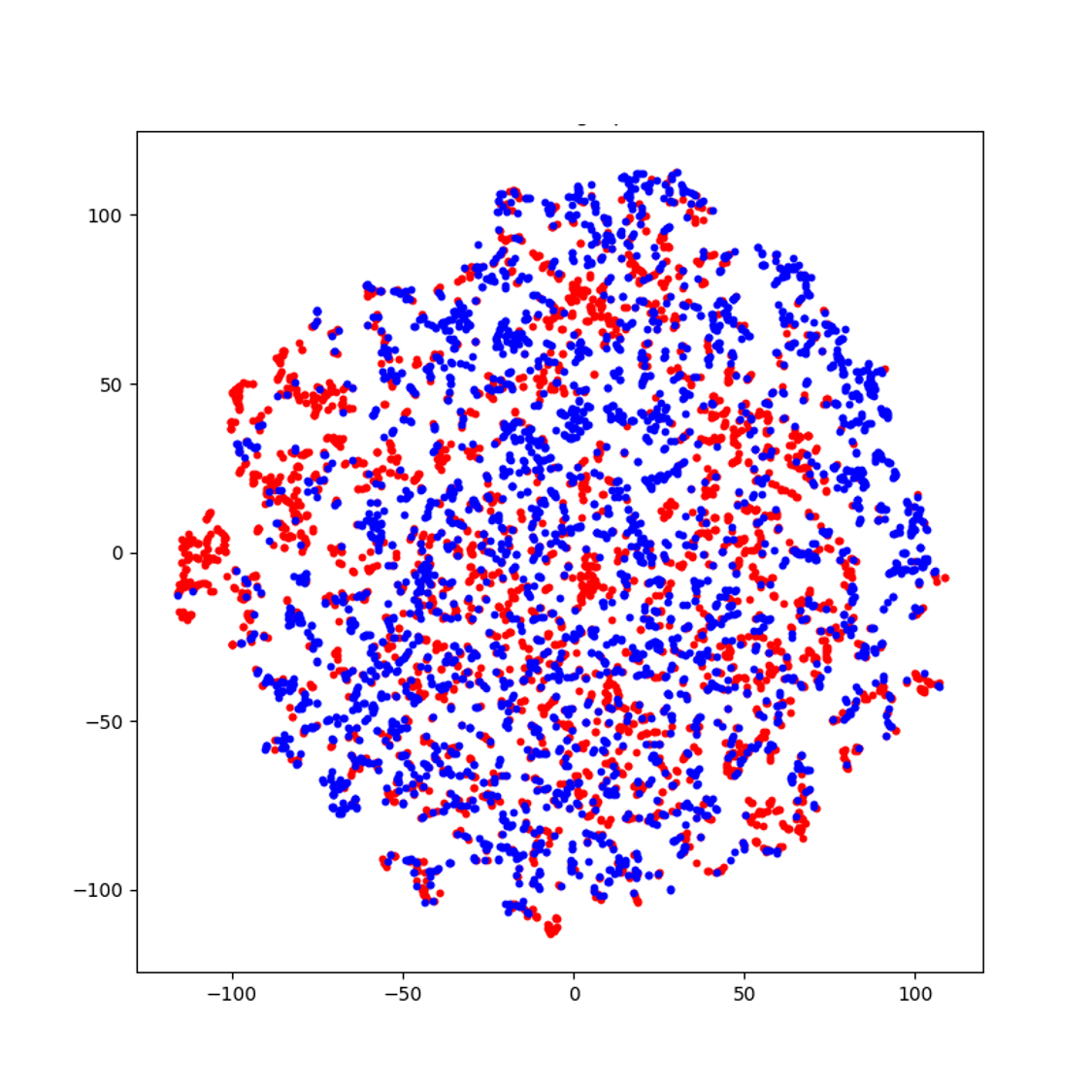}
%\caption{no_dann_demographics}
\end{minipage}
}%
\subfigure[Adapted migration]{
\begin{minipage}[t]{0.24\linewidth}
\centering
\includegraphics[width=1.8in]{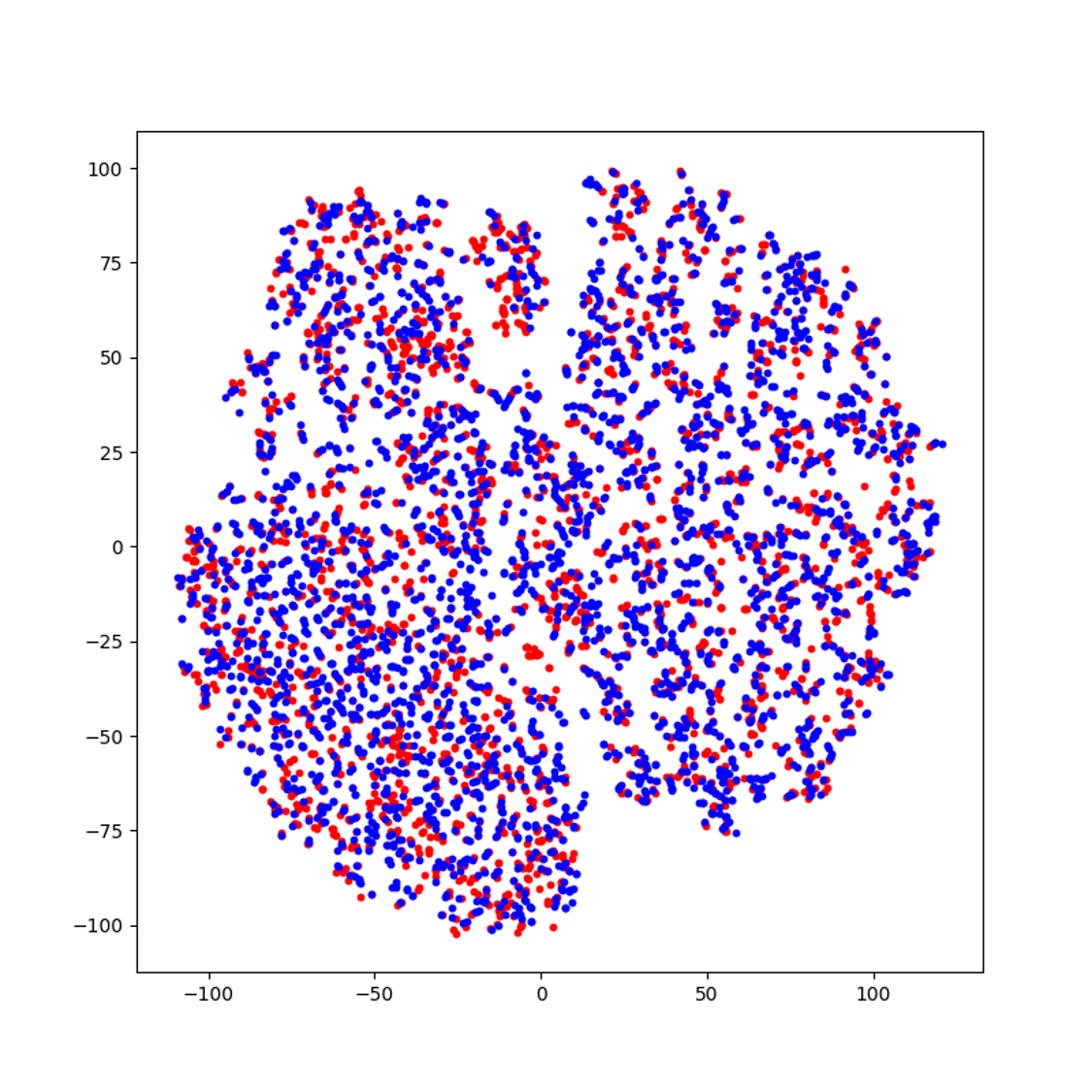}
%\caption{dann_demographics}
\end{minipage}
}%
\caption{The effect of FG-based domain adaptation on the distribution of learned feature representation (best viewed in color). (a)-(d) and (e)-(h) show the the t-SNE embeddings of the feature representations learned by $\text{PrADA}_{\text{w/o DA\&IR}}$ and $\text{PrADA}_{\text{w/o IR}}$, respectively, on feature groups of Census Income data. Red points correspond to the source domain examples, while blue ones correspond to the target domain samples. In all feature groups, the adaptation makes the two distributions of learned feature representations much closer.}
\label{tse_da}
\end{figure*}

\subsection{Experiments on Census Income}

\noindent\textbf{Census Income} is a census dataset from the UCI Machine Learning Repository. We split this dataset into a $\textit{undergraduate}$ source domain and a $\textit{postgraduate}$ target domain. The source domain has 80000 labeled examples, while the target domain has 4000 labeled samples and 9000 unlabeled samples. Our goal is to help party A of the target domain to predict whether a person’s income exceeds 50,000 US dollars or not. 

After data preprocessing, the census income dataset contains 36 features, 31 of which are categorical. We put 5 numerical features on active parties (i.e., A and B) while the 31 categorical features on passive party C. We split 31 features on party C into 4 feature groups (FG) including \textit{employment(emp)}, \textit{demographics(demo)}, \textit{household(house)}, and \textit{migration(migr)}. Thus, we have $C^4_2$(i.e., 6) interactive feature groups, which are \textit{emp-demo}, \textit{emp-house}, \textit{emp-migr}, \textit{demo-house}, \textit{demo-migr}, \textit{house-migr}. We embed all categorical features into dense vectors. Table \ref{fe_census} shows the architecture of feature extractor for each of the 10 feature groups and the one (i.e., \textit{all\_feat}) for all features when feature grouping is not applied.

% the LR model has parameter dimension of 9 (5+4).

\begin{table}[tb]
    \caption{Architecture of feature extractors for Census Income dataset. All feature extractors only use fully-connected layers, and adopt Leaky ReLU as the activation function, which is omitted in the table for simplicity.}
    \label{fe_census}
    \centering
\begin{tabular}{l||c}
\hline
\multicolumn{1}{c}{FG name} & \multicolumn{1}{l}{feature extractor architecture} \\
\hline
\hline
 emp & FC(28$\rightarrow$56)-FC(56$\rightarrow$28)-FC(28$\rightarrow$14) \\
 demo & FC(25$\rightarrow$50)-FC(50$\rightarrow$25)-FC(25$\rightarrow$12) \\
 migr & FC(56$\rightarrow$86)-FC(86$\rightarrow$56)-FC(56$\rightarrow$18) \\
 house & FC(27$\rightarrow$54)-FC(54$\rightarrow$27)-FC(27$\rightarrow$13) \\
 emp-demo & FC(53$\rightarrow$78)-FC(78$\rightarrow$53)-FC(53$\rightarrow$15) \\
 emp-migr & FC(84$\rightarrow$120)-FC(120$\rightarrow$84)-FC(84$\rightarrow$20) \\
 emp-house & FC(51$\rightarrow$81)-FC(81$\rightarrow$55)-FC(55$\rightarrow$15) \\
 demo-migr & FC(81$\rightarrow$120)-FC(120$\rightarrow$81)-FC(81$\rightarrow$20) \\
 demo-house & FC(52$\rightarrow$78)-FC(78$\rightarrow$52)-FC(52$\rightarrow$15) \\
 migr-house & FC(83$\rightarrow$120)-FC(120$\rightarrow$83)-FC(83$\rightarrow$20) \\
 all\_feat & FC(136$\rightarrow$150)-FC(150$\rightarrow$60)-FC(60$\rightarrow$20) \\
\hline

\end{tabular}
\end{table}

\begin{table*}[th]
    \caption{Comparison between models in different settings on Loan Default dataset.}
    \label{exp_ppd}
    \centering
\begin{tabular}{l|l||c|c||c|c||c|c}
\hline
\multicolumn{2}{c}{Positive labels} & \multicolumn{2}{c}{40} & \multicolumn{2}{c}{80} & \multicolumn{2}{c}{160} \\
\hline
\multicolumn{1}{c}{Setting} & \multicolumn{1}{l}{Model} & \multicolumn{1}{c}{AUC (\%)} & \multicolumn{1}{c}{KS (\%)} & \multicolumn{1}{c}{AUC (\%)} & \multicolumn{1}{c}{KS (\%)} & \multicolumn{1}{c}{AUC (\%)} & \multicolumn{1}{c}{KS (\%)} \\
\hline
\hline
 $\mathbf{A}$-$\mathbf{Local}$ & LR & 57.17 & 12.65 & 56.51 & 13.16 & 57.77 & 15.14\\
    & XGBoost & 56.66 & 11.49 & 57.90 & 14.89 & 58.91 & 17.47 \\
\hline
 $\mathbf{A}$-$\mathbf{VFL}$ & SecureLR & 59.67 & 15.31 & 64.68 & 24.10  & 67.78 & 28.61 \\
     & SecureBoost & 57.88 & 12.86 & 64.90 & 23.33  & 70.68 & 31.09 \\
     & $\text{PrADA}_{\text{w/o DA\&FG\&IR}}$ & 63.26$\pm$0.88 & 21.14$\pm$1.26 & 67.49$\pm$0.78 & 28.06$\pm$1.12 & 68.66$\pm$0.94 & 29.59$\pm$1.39 \\
\hline
$\mathbf{AB}$-$\mathbf{VFL}$ & SecureLR & 72.72 & 35.81 & 73.04 & 35.90 & 74.22 & 36.87\\
      & SecureBoost & 75.18 & 38.53 & 75.96 & 40.65 & 76.16 & 41.83\\
     & $\text{PrADA}_{\text{w/o DA\&FG\&IR}}$ & 75.11$\pm$0.37 & 40.28$\pm$1.03 & 75.16$\pm$0.19 & 40.63$\pm$0.64 & 75.53$\pm$0.23 & 41.12$\pm$0.71 \\
\hline
\hline
% $\checkmark$ & $\checkmark$ & $\text{PrADA}_{\text{w/o FG}}$ & 80.01 & 46.65 \\
 $\mathbf{B} \rightarrow \mathbf{A}$ & $\text{PrADA}_{\text{w/o FG\&IR}}$ & 75.27$\pm$0.25 & 40.82$\pm$0.53 & 75.52$\pm$0.22 & 41.25$\pm$0.24 & 75.76$\pm$0.28 & 41.91$\pm$0.56 \\
    & $\text{PrADA}_{\text{w/o IR}}$ & 75.63$\pm$0.11 & 41.42$\pm$0.74  & 75.84$\pm$0.09 & 42.04$\pm$0.61 & 76.43$\pm$0.08 & 42.61$\pm$0.17   \\
    & $\text{PrADA}$ & \textbf{75.75$\pm$0.12} & \textbf{41.69$\pm$0.36}  & \textbf{75.99$\pm$0.05} & \textbf{42.48$\pm$0.21} & \textbf{76.58$\pm$0.18} & \textbf{43.48$\pm$0.62}   \\
\hline
\end{tabular}
\end{table*}

The experimental results are shown in Table \ref{exp_census}. From these results, we observe that: (1) SecureBoost and SecureLR in $\mathbf{A}$-$\mathbf{VFL}$ outperform their counterparts in $\mathbf{A}$-$\mathbf{Local}$ demonstrating that leveraging additional features improve the model performance. (2) The performance of models in $\mathbf{B} \rightarrow \mathbf{A}$ significantly outperforms that of models in $\mathbf{A}$-$\mathbf{VFL}$ across all settings of different positive labels. This is expected because a considerable amount of source data is involved in training. More specifically, when the number of positive labels is small (i.e., 40), the performance gain is the most significant. For example, PrADA outperforms $\text{PrADA}_{\text{w/o DA\&FG\&IR}}$ in AUC by 5.45\% and in KS by 9.56\%, and outperforms SecureBoost in AUC by 7.2\% and in KS by 10.19\% when the positive label is 40. (3) $\text{PrADA}_{\text{w/o FG\&IR}}$ in $\mathbf{B} \rightarrow \mathbf{A}$ outperforms $\text{PrADA}_{\text{w/o DA\&FG\&IR}}$ in $\mathbf{AB}$-$\mathbf{VFL}$ in AUC by 1.02\% and in KS by 0.60\% on average, and outperforms SecureBoost in AUC by 0.60\% and in KS by 0.94\% on average, demonstrating the effectiveness of PrADA on bridging the divergence between source and target domains. 
(4) In $\mathbf{B} \rightarrow \mathbf{A}$ setting, $\text{PrADA}_{\text{w/o IR}}$ outperforms $\text{PrADA}_{\text{w/o FG\&IR}}$ in AUC by 0.17\% and KS by 0.58\% on average, demonstrating the effectiveness of FG-based domain adversarial training on improving the transferability of feature extractors. In addition to boosting model performance, feature grouping also enhances the interpretability of target model $R^A$, which we discuss in section \ref{interpret}. (5) In $\mathbf{B} \rightarrow \mathbf{A}$ setting, $\text{PrADA}$ outperforms $\text{PrADA}_{\text{w/o IR}}$ in AUC by 0.31\% and KS by 0.70\% on average, demonstrating the interaction on feature groups help improve model performance.

To dive deeper into the effect of FG-based domain adaptation on learned feature representations, we visualize the t-SNE embeddings~\cite{pmlr-v32-donahue14} of the feature representations in Figure \ref{tse_da}. Figure \ref{tse_da}(a)-(d) and Figure \ref{tse_da}(e)-(h) show the the t-SNE embeddings of the feature representations learned by $\text{PrADA}_{\text{w/o DA\&IR}}$ and $\text{PrADA}_{\text{w/o IR}}$, respectively, on feature groups of Census Income data. We observe that the adaptation in our method makes the two distributions of learned feature representations much closer in all feature groups.

\subsection{Experiments on Loan Default}

\noindent\textbf{Loan Default} is a loan default risk dataset for the online lending industry published by FinVolution Group. It contains loan data issued in 2014. We consider 40000 labeled samples of loans issued in the first three quarters of 2014 as the source domain while the 4000 labeled samples and 9000 unlabeled samples of loans issued in the fourth quarter as the target domain. This is an Out-Of-Time scenario in financial risk control. Our goal is to help party A to build a loan predictor to predict whether a loan will default or not.

After data preprocessing, the Load Default dataset has 162 features, 27 of which are categorical. For protecting privacy, user and feature names are anonymized. We put 6 \textit{demographics} features and labels on active parties, while the rest 156 features on passive party C. We split features of party C into 5 groups including \textit{user location(loc)}, \textit{third-party period(period)}, \textit{education(edu)}, \textit{social network (soc)}, and \textit{micro-blog(mblog)}. Thus, we have $C^5_2$(i.e., 10) interactive feature groups, which are \textit{loc-period}, \textit{loc-edu}, \textit{loc-soc}, \textit{loc-mblog}, \textit{period-edu}, \textit{period-soc}, \textit{period-mblog}, \textit{edu-soc}, \textit{edu-mblog}, \textit{soc-mblog}. We embed all categorical features into dense vectors. Table \ref{fe_ppd} shows the architecture of feature extractor for each of the 15 feature groups and the one (i.e., \textit{all\_feat}) for all features when feature grouping is not applied.

% The LR model has parameter dimension of 11 (6+5)

\begin{table}[tb]
    \caption{Architecture of feature extractors for Loan Default dataset. All feature extractors only use fully-connected layers, and adopt Leaky ReLU as the activation function, which is omitted in the table for simplicity.}
    \label{fe_ppd}
    \centering
\begin{tabular}{l||c}
\hline
\multicolumn{1}{c}{FG name} & \multicolumn{1}{l}{feature extractor architecture} \\
\hline
\hline
 loc & FC(15$\rightarrow$20)-FC(20$\rightarrow$15)-FC(15$\rightarrow$6) \\
 period & FC(85$\rightarrow$100)-FC(100$\rightarrow$60)-FC(60$\rightarrow$8) \\
 edu & FC(30$\rightarrow$50)-FC(50$\rightarrow$30)-FC(30$\rightarrow$6) \\
 soc & FC(18$\rightarrow$30)-FC(30$\rightarrow$18)-FC(18$\rightarrow$6) \\
 mblog & FC(55$\rightarrow$70)-FC(70$\rightarrow$30)-FC(30$\rightarrow$8) \\
 loc-period & FC(100$\rightarrow$120)-FC(120$\rightarrow$75)-FC(75$\rightarrow$14) \\
 loc-edu & FC(45$\rightarrow$70)-FC(70$\rightarrow$45)-FC(45$\rightarrow$12) \\
 loc-soc & FC(33$\rightarrow$50)-FC(50$\rightarrow$33)-FC(33$\rightarrow$12) \\
 loc-mblog & FC(70$\rightarrow$90)-FC(90$\rightarrow$45)-FC(45$\rightarrow$14) \\
 period-edu & FC(115$\rightarrow$150)-FC(150$\rightarrow$90)-FC(90$\rightarrow$14) \\
 period-soc & FC(103$\rightarrow$130)-FC(130$\rightarrow$78)-FC(78$\rightarrow$14) \\
 period-mblog & FC(140$\rightarrow$170)-FC(170$\rightarrow$90)-FC(90$\rightarrow$16) \\
 edu-soc & FC(48$\rightarrow$80)-FC(80$\rightarrow$48)-FC(48$\rightarrow$12) \\
 edu-mblog & FC(85$\rightarrow$120)-FC(120$\rightarrow$60)-FC(60$\rightarrow$14) \\
 soc-mblog & FC(73$\rightarrow$100)-FC(100$\rightarrow$48)-FC(48$\rightarrow$14) \\
 all\_feat & FC(203$\rightarrow$210)-FC(210$\rightarrow$70)-FC(70$\rightarrow$20) \\
\hline

\end{tabular}
\end{table}

% The experimental results are shown in Table 1. From the results, we observe that: (1) the domain divergence is considerable since directly combining data of source and target domains would overall degrade the performance of NDA models; (2) the performance of DANN is better than or comparable to that of NDA models trained on all data, which is somewhat expected since DANN diminishes domain divergence; (3) our model achieves the best AUC demonstrating the effectiveness of the combination of domain adaptation and feature grouping.

The experimental results are reported in Table \ref{exp_ppd}. From these, we observe that: 
% (1) The performance of models trained in $\mathbf{A}$-$\mathbf{Local}$ setting is only slightly better than that of random guessing, indicating that party A is unable to build a useful model using only local data. 
% This is somewhat expected since the active parties only have 4 features locally, which is insufficient to build a useful loan default predictor or conduct domain adaptation. 
(1) Table \ref{exp_ppd} reports a similar trend as Table \ref{exp_census} does that the performance of models improves from $\mathbf{A}$-$\mathbf{Local}$ setting to $\mathbf{A}$-$\mathbf{VFL}$ and then to $\mathbf{B} \rightarrow \mathbf{A}$ as more data is involved in training. The performance gains the most when the number of positive labels is small. Specifically, when the number of positive labels is 40, PrADA in $\mathbf{B} \rightarrow \mathbf{A}$ setting outperforms $\text{PrADA}_{\text{w/o DA\&FG\&IR}}$ in $\mathbf{A}$-$\mathbf{VFL}$ setting in AUC and KS by 12.49\% and 20.55\% respectively, and outperforms SecureBoost in AUC and KS by 17.87\% and 28.83\% respectively. As the number of positive labels increases, the performance gain narrows. (2) $\text{PrADA}_{\text{w/o FG\&IR}}$ in $\mathbf{B} \rightarrow \mathbf{A}$ setting outperforms $\text{PrADA}_{\text{w/o DA\&FG\&IR}}$ in $\mathbf{AB}$-$\mathbf{VFL}$ in AUC by 0.25\% and in KS by 0.65\% on average, demonstrating the effectiveness of PrADA on mitigating domain divergence. (3) In $\mathbf{B} \rightarrow \mathbf{A}$ setting, $\text{PrADA}_{\text{w/o IR}}$ outperforms $\text{PrADA}_{\text{w/o FG\&IR}}$, demonstrating the superiority of FG-based DA over conventional DA, and $\text{PrADA}$ outperforms $\text{PrADA}_{\text{w/o IR}}$, demonstrating the interaction on feature groups help enhance model performance. 

\subsection{Model Interpretability}
\label{interpret}
% In addition to improving the performance of target label predictor $R^B$, $\text{PrADA}$ enhances the explainability of $R^B$. Take the $R^B$ trained on Tencent data as an example. It takes as input 13 interpretable high-order features, instead of low-interpretable raw features or feature representation.  

% \begin{table}[tb!]
%     \centering
% \begin{tabular}{l|l||c|c}
% \hline
% \multicolumn{1}{c}{Setting} & \multicolumn{1}{l}{Model} & \multicolumn{1}{c}{AUC (\%)} & \multicolumn{1}{c}{KS (\%)} \\
% \hline
% \hline
%  $\mathbf{A}$-$\mathbf{Local}$ & LR & 54.34 & 11.10 \\
%     & XGBoost & 54.29 & 9.34 \\
% \hline
%  $\mathbf{A}$-$\mathbf{VFL}$ & SecureLR & 62.97 & 22.43 \\
%      & SecureBoost & 65.78 & 24.79 \\
%      & $\text{PrADA}_{\text{w/o DA}}$ & 64.85 & 26.02 \\
% \hline
% % \hline
% $\mathbf{AB}$-$\mathbf{VFL}$ & SecureLR & 72.55 & 39.28 \\
%       & SecureBoost & 74.69 & 41.66 \\
%      & $\text{PrADA}_{\text{w/o DA}}$ & 75.03$\pm$0.40 & 41.78$\pm$0.41 \\
% \hline
% \hline
%  $\mathbf{B} \rightarrow \mathbf{A}$
%  & $\text{PrADA}_{\text{w/o FG}}$ & 75.21$\pm$0.31 & 41.04$\pm$0.26 \\
%     & $\text{PrADA}$ & 75.33$\pm$0.18 & 42.56$\pm$0.35  \\
%     % & $\text{PrADA}_{ft}$ & 75.58 &  42.75 \\
% \hline
% \end{tabular}
%     \caption{Comparison between models on loan default dataset.}
%     \label{tencent}
% \end{table}

We demonstrate model interpretability by visualizing the impact of features on target model $R^A$ using SHAP~\cite{NIPS2017_7062}, a tool widely used to explain black-box models. As discussed in section \ref{sfl}, the real values of model parameters of $R^A$ are not accessible by either party C or party A. This means that party A cannot interpret the model by simply looking at model parameters. SHAP provides party A with a way to interpret the model without accessing the model parameters. We select the Census Income dataset for this purpose since the semantic meaning of features in the Loan Default dataset is anonymized.

% Since LR maps its input to a probability between 0 and 1, the weights ${W}$ of LR do not influence the probability linearly. Here, we adopt $\textit{odds ratio}$ $e^{\mathbf{W}_j}$ \cite{molnar2019} to represent the weight of feature $\mathbf{x}_j$. The term $\textit{odds}=\frac{p(y=1)}{p(y=0)}=e^{\mathbf{Wx}}$ indicates the probability of event divided by the probability of no event. Thus, in loan default prediction, the $\textit{odds ratio}=e^{\mathbf{W}_j(\mathbf{x}_j+1)-\mathbf{W}_j\mathbf{x}_j}=e^{\mathbf{W}_j}$ of feature $\mathbf{x}_j$ can be interpreted as a change in  $\mathbf{x}_j$ by one unit changes the odds of default v.s. no default by a factor of $e^{\mathbf{W}_j}$, given all other feature weights remain fixed. Figure \ref{fi} depicts the boxplot showing the statistical summary of odds ratios of the 13 features.
% \begin{figure}[h]
% \centering
%     \subfigure[]{\includegraphics[width=0.48\linewidth]{figs/fi_census.png} \label{fi_on_ce}}
%     \subfigure[]{\includegraphics[width=0.48\linewidth]{figs/fi_loan_default.png} \label{fi_on_ld}} 
%     \caption{Feature importance}
%     \label{fi}
% \end{figure}

% \begin{figure}[ht]
% \centering
%     \includegraphics[width=8cm]{figs/boxplot.png}
%     \caption{boxplot showing minimum, first quartile, median, third quartile, and maximum of odds ratios of 13 features.}
%     \label{fi}
% \end{figure}

\begin{figure}[ht]
\centering
    \includegraphics[width=8cm]{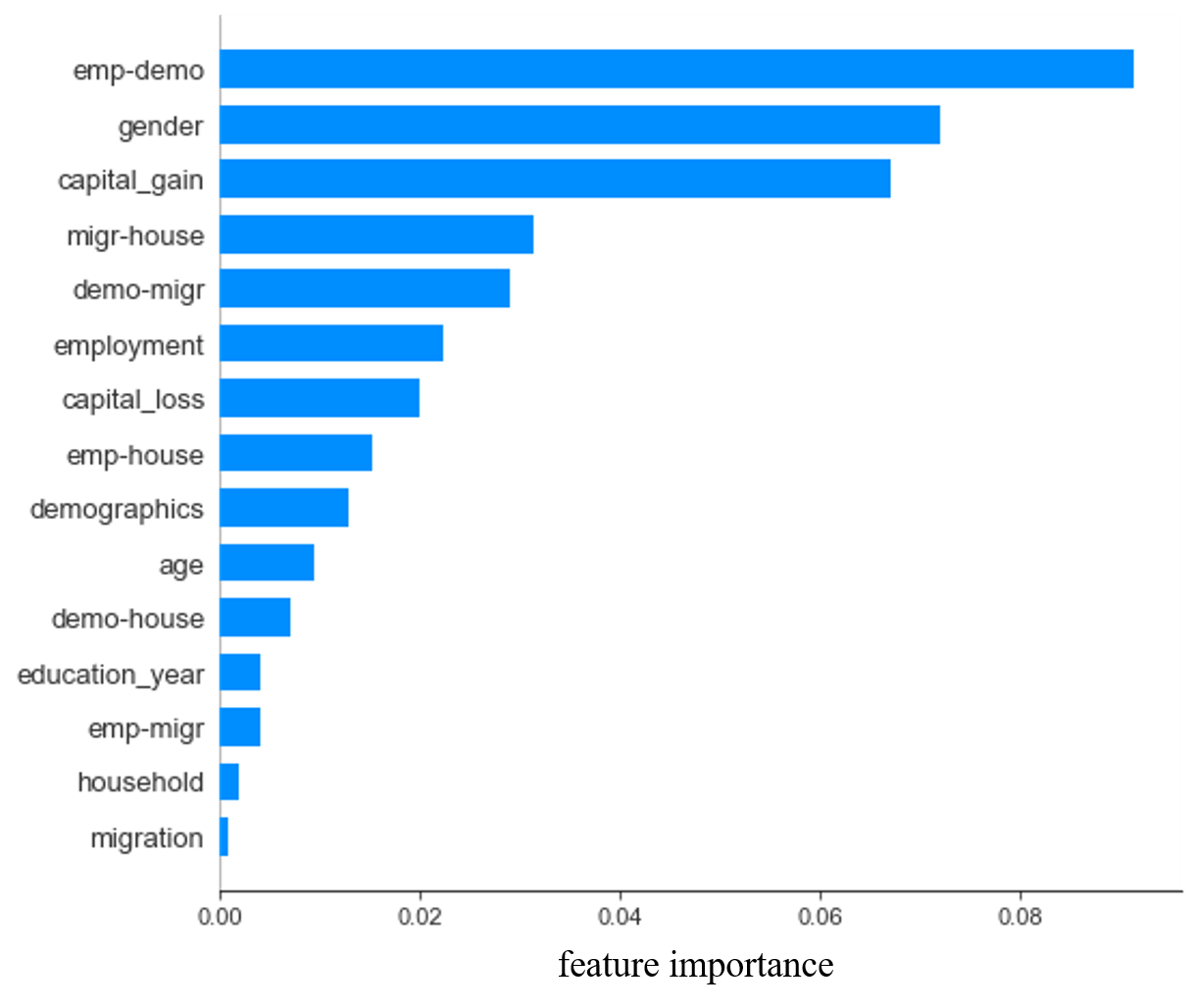}
    \caption{The importance of features produced by SHAP. Each feature can be a raw feature from party A, or a (interactive) feature group from party C.}
    \label{feat_imp}
\end{figure}

Figure \ref{feat_imp} lists the most influential features of model $R^A$ in descending order. The top features have higher predictive power because they contribute more to the model than the bottom ones. For example, \textit{emp-demo}, \textit{gender}, \textit{capital\_gain}, \textit{migr-house} and  \textit{demo-migr} are the top-5 most influential features.

\begin{figure}[ht]
\centering
    \includegraphics[width=8cm]{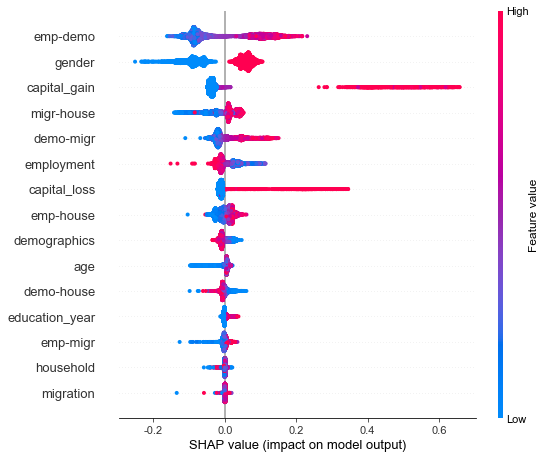}
    \caption{The impact of features on model predictions. Each feature can be a raw feature from the party A, or a high-level feature representing a (interactive) feature group from the party C.}
    \label{shap_value}
\end{figure}

The SHAP can further show the positive and negative relationships of features with the prediction target. Figure \ref{shap_value} plots the SHAP values of every feature for all samples to illustrate the impact of those features on the prediction output. Features are ranked in descending order. The color represents the feature value (red high, blue low). The horizontal location shows whether the effect of a feature value is associated with a higher or lower prediction. Specifically, \textit{emp-demo},  \textit{captial\_gain} and \textit{demo-migr} are positively correlated with the prediction, while \textit{gender} and \textit{migr-house} are negatively correlated with the prediction.

% while the other four features have positive impact on the label prediction.

% We use SHAP, which is a approach that is widely used to explain various black-box models, to explain the influence of each feature on the label predictor model, as shown in Figure \ref{shap_value}. Each row of firgure \ref{shap_value} represents a feature group, and the abscissa denotes the SHAP value. The redder the color indicates the larger the value of the feature group itself, and the bluer the color indicates the smaller the value of the feature group itself. It intuitively shows that employment is a very vital feature group and is positively correlated with income. The stock\_dividends feature group will also significantly affects income, and a high stock\_dividends value will greatly increase the prediction of income.

% SHAP is a machine learning library widely used to explain various black-box models (including XGBoost and neural networks). We use SHAP to explain the influence of each feature group in Prada on the predicted value of each sample. Each row of firgure \ref{shap_value} represents a feature group, and the abscissa denotes the SHAP value. The redder the color indicates the larger the value of the feature group itself, and the bluer the color indicates the smaller the value of the feature group itself. It intuitively shows that employment is a very vital feature group and is positively correlated with income. The stock\_dividends feature group will also significantly affects income, and a high stock\_dividends value will greatly increase the prediction of income.

\subsection{Computation Cost}
We compare the training time among SecureLR, SecureBoost, and $\text{PrADA}$ using FATE 1.6. The experiments are conducted on a machine with 72 Intel Xeon Gold 6140 CPUs and 320 GB RAM. All experiments are simulated in standalone deployment mode. Note that the privacy-preserving VFL framework (PP-VFL) discussed in section \ref{sfl} has been integrated into FATE, while the federated adversarial domain adaptation (FADA) discussed in section \ref{fada} has not because some of the core functionalities of FADA are not satisfied by FATE for now. Thus, we estimate the training time of PrADA by simulation using PP-VFL on FATE. 

\begin{table}[ht!]
    \caption{Training time (hours) on Census Income dataset. PT and FT denote pre-taining time and fine-tuning time, respectively.}
    \label{tencent_time}
    \centering
\begin{tabular}{l|l||c|c}
\hline
\multicolumn{1}{c}{Setting} & \multicolumn{1}{l}{Model} & \multicolumn{1}{c}{Time(h)} & \multicolumn{1}{c}{}\\
\hline
\hline
%  $\mathbf{A}$-$\mathbf{VFL}$ & SecureLR & 196 \\
%      & SecureBoost & 312 \\
%      & $\text{PrADA}_{\text{w/o DA}}$ & - \\
% \hline
% \hline
$\mathbf{AB}$-$\mathbf{VFL}$ & SecureLR & $\sim$1.12 \\
      & SecureBoost & $\sim$2.16 \\
     & $\text{PrADA}_{\text{w/o DA\&FG\&IR}}$ & $\sim$4.10 \\
\hline
\hline
\multicolumn{1}{c}{} & \multicolumn{1}{l}{} & \multicolumn{1}{c}{PT(h)} & \multicolumn{1}{c}{FT(h)}\\
\hline
 $\mathbf{B} \rightarrow \mathbf{A}$ & $\text{PrADA}_{\text{w/o IR}}$ & $\sim$5.47  & $\sim$0.52 \\
  & $\text{PrADA}$ & $\sim$8.24  & $\sim$1.32 \\
\hline
\end{tabular}
\end{table}

Table \ref{tencent_time} reports the training time of $\text{PrADA}_{\text{w/o DA\&FG\&IR}}$ is roughly 4.10 hours, which is approximately twice the training time spent by SecureBoost. In $\mathbf{B} \rightarrow \mathbf{A}$ setting, $\text{PrADA}_{\text{w/o IR}}$ takes 5.47 hours to train because FG-based domain adaptation is involved. However, once pre-training is completed, $\text{PrADA}_{\text{w/o IR}}$ only takes half an hour to perform fine-tuning. $\text{PrADA}$ takes 8.24 hours to train because it spends extra time on feature group interaction, additional feature extractor training and feature representations encryption. As reported in Table \ref{exp_census} and Table \ref{exp_ppd}, $\text{PrADA}$ exceeds $\text{PrADA}_{\text{w/o IR}}$ by only a small margin. Therefore, if efficiency is a major concern, $\text{PrADA}_{\text{w/o IR}}$ is a better choice.
% \section{Conclusions and Future Work}\label{sec:experiments}

\section{Conclusion}
In this paper, we propose a privacy-preserving vertical federated adversarial domain adaptation approach.  In particular, we develop a privacy-preserving VFL framework that allows participating parties to collaboratively conduct domain adaptation without exposing private data. To reduce feature dimensionality, enhance model interpretability, and facilitate the learning of domain-invariant features, we propose a fine-grained adversarial domain adaptation over feature groups that each holds tightly relevant features. Experiments demonstrate both the effectiveness and practicality of our approach.

% \section{Acknowledge}
% This work is partially supported by the National Key Research and Development Program of China under grant [2018AAA0101100].

\ifCLASSOPTIONcompsoc
  \section*{Acknowledgments}
\else
  \section*{Acknowledgment}
\fi

This work is partially supported by the National Key Research and Development Program of China under grant [2018AAA0101100].
%{\color{red}The authors would like to thank...}

\ifCLASSOPTIONcaptionsoff
  \newpage
\fi

\bibliographystyle{plain}
\bibliography{prada}


% \end{IEEEbiography}


\begin{IEEEbiographynophoto}{Yan Kang}
is currently a research team lead with the AI department of WeBank, Shenzhen, China. His works focus on the research and implementation of privacy-preserving machine learning and federated learning. His research was authored or coauthored in well-known conferences and journals including IEEE Intelligence Systems, IJCAI, and ACM TIST, and coauthored the Federated Learning book.
\end{IEEEbiographynophoto}
\begin{IEEEbiographynophoto}{Yuanqin He} is currently a researcher with WeBank. He received the B.S. degree from Shanghai Jiao Tong University, and the Ph.D. degree in Physics from Technical University of Munich. His research interests include machine learning and federated learning.
\end{IEEEbiographynophoto}
\begin{IEEEbiographynophoto}{Jiahuan Luo} is currently a researcher with WeBank. He received the B.S. degree from Guangdong University of Foreign Studies and the Master degree in Software Engineering from South China University of Technology. His research interests include federated learning and representation learning.
\end{IEEEbiographynophoto}
\begin{IEEEbiographynophoto}{Yang Liu}
is an associate professor with institute for AI Industry Research (AIR), Tsinghua University. Her research interests include federated learning, machine learning, multi-agent systems, statistical mechanics and AI industrial applications. Her research work was recognized with multiple awards, such as AAAI Innovation Award and CCF Technology Award.
\end{IEEEbiographynophoto}
\begin{IEEEbiographynophoto}{Tao Fan}
is a tech lead with the AI department of WeBank, ShenZhen, China. He is now responsible for FATE, an industrial level federated learning open source project. He has more than 8 years of experience in large-scale machine learning. He received his Master degree from University of Science and Technology of China in 2013. 
\end{IEEEbiographynophoto}
\begin{IEEEbiographynophoto}{Qiang Yang}
is a fellow of Royal Society of Canada (RSC) and Canadian Academy of Engineering (CAE), Chief Artificial Intelligence Officer of WeBank, a Chair Professor of Computer Science and Engineering Department at Hong Kong University of Science and Technology (HKUST). 
% He is the Conference Chair of AAAI-21, the Honorary Vice President of Chinese Association for Artificial Intelligence(CAAI) , the President of Hong Kong Society of Artificial Intelligence and Robotics(HKSAIR) and the President of Investment Technology League (ITL), Independent Non-executive Director of China Mobile Limited. 
He is a fellow of AAAI, ACM, CAAI, IEEE, IAPR, AAAS. 
% He was the Founding Editor in Chief of the ACM Transactions on Intelligent Systems and Technology (ACM TIST) and the Founding Editor in Chief of IEEE Transactions on Big Data (IEEE TBD). He received the ACM SIGKDD Distinguished Service Award in 2017 and the Wu Wenjun outstanding contribution award of artificial intelligence in 2019. He graduated from Peking University. He received his PhD from the Department of Computer Science at the University of Maryland, College Park in 1989 and had been a faculty member at the University of Waterloo between 1989 and 1995. 
His research interests are artificial intelligence, machine learning, data mining and planning. 
% He had been the Founding Director of the Huawei's Noah's Ark Research Lab between 2012 and 2015, the Founding Director of HKUST’s Big Data Institute, the Founder of 4Paradigm and the President of IJCAI (2017-2019). 
His latest books are Transfer Learning, Federated Learning and Practicing Federated Learning.
\end{IEEEbiographynophoto}

%\newpage



% \vfill

% \enlargethispage{-5in}

\end{document}